\documentclass[10pt,twocolumn,letterpaper]{article}

\usepackage[pagenumbers]{cvpr} 

\usepackage{multirow}
\usepackage[pdftex]{graphicx}
\usepackage{amsmath}
\usepackage{amssymb}
\usepackage{booktabs}
\usepackage{xcolor}

\usepackage[pagebackref,breaklinks,colorlinks]{hyperref}

\usepackage[capitalize]{cleveref}
\crefname{section}{Sec.}{Secs.}
\Crefname{section}{Section}{Sections}
\Crefname{table}{Table}{Tables}
\crefname{table}{Tab.}{Tabs.}

\newcommand{\figref}[1]{Fig.~\ref{#1}}
\newcommand{\tabref}[1]{Tab.~\ref{#1}}

\newcommand{\secref}[1]{Sec.~\ref{#1}}

\def\eg{\emph{e.g.~}} 
\def\ie{\emph{i.e.~}} 
 
\def\etc{\emph{etc.~}} \def\vs{\emph{vs.~}}
\def\wrt{w.r.t.~} 
 
\def\etal{\emph{et al.~}}

\newcommand{\vlnbert}{VLN$\circlearrowright$BERT }

\def\cvprPaperID{12664} 
\def\confName{CVPR}
\def\confYear{2023}

\begin{document}


    \title{Predicting Topological Maps for Visual Navigation in Unexplored Environments}
\author{
Huangying Zhan\thanks{This work was done when HZ was with The University of Adelaide}\\
InnoPeak Technology, Inc.\\
{\tt\small zhanhuangying.work@gmail.com}
\and
Hamid Rezatofighi \\
Monash University\\
{\tt\small Hamid.Rezatofighi@monash.edu}
\and
Ian Reid \\
The University of Adelaide\\
{\tt\small ian.reid@adelaide.edu.au}
}
\maketitle

\begin{abstract}
We propose a robotic learning system for autonomous exploration and navigation in unexplored environments. We are motivated by the idea that even an unseen environment may be familiar from previous experiences in similar environments.    
The core of our method, therefore, is a process for building, predicting, and using probabilistic layout graphs for assisting goal-based visual navigation.
We describe a navigation system that uses the layout predictions to satisfy high-level goals (\eg ``go to the kitchen") more rapidly and accurately than the prior art.
Our proposed navigation framework comprises three stages: 
(1) \textit{Perception and Mapping}: 
building a multi-level 3D scene graph;
(2) \textit{Prediction}: 
predicting probabilistic 3D scene graph for the unexplored environment;
(3) \textit{Navigation}: 
assisting navigation with the graphs.
We test our framework in Matterport3D and show more success and efficient navigation in unseen environments. 
\end{abstract}
    
    \section{Introduction}
\label{sec:intro}
Human navigation in an unknown environment can rely on a variety of information in addition to our immediate percepts: a map, instructions from others who have visited the environment before, or prior knowledge of similar environmental structures and layouts. It is this latter competence -- the ability to use ``familiarity'' (patterns learnt from prior experience in similar but not identical environments) -- that is the main idea we seek to exploit in the present work. Specifically, we aim to leverage the prior experience of navigating in a similar indoor environment to improve an agent's ability to explore or purposefully navigate in a new, previously unseen building.

Although the exact representations used by humans for navigation are unknown, it is clear that for the most part, we encode spatial information using high-level -- often topological -- abstractions rather than using detailed geometry. 
Likewise, navigation actions are accomplished without recourse to detailed global geometry, but through a combination of local control 
and goal-directed behaviour and exploration.
In aiming to endow robots with similar high-level capabilities for visual navigation in an \textit{unknown} but \textit{familiar} environment, we therefore employ a topological representation -- 3D Scene Graph -- to encode the environment structure. We consider the problem of navigating to a user-specified goal in a previously unseen environment, and develop methods that enable the agent to make informed predictions about its actions that lead to faster exploration and goal-satisfaction.

\begin{figure*}[t!]
        \centering
        \includegraphics[width=0.7\textwidth]{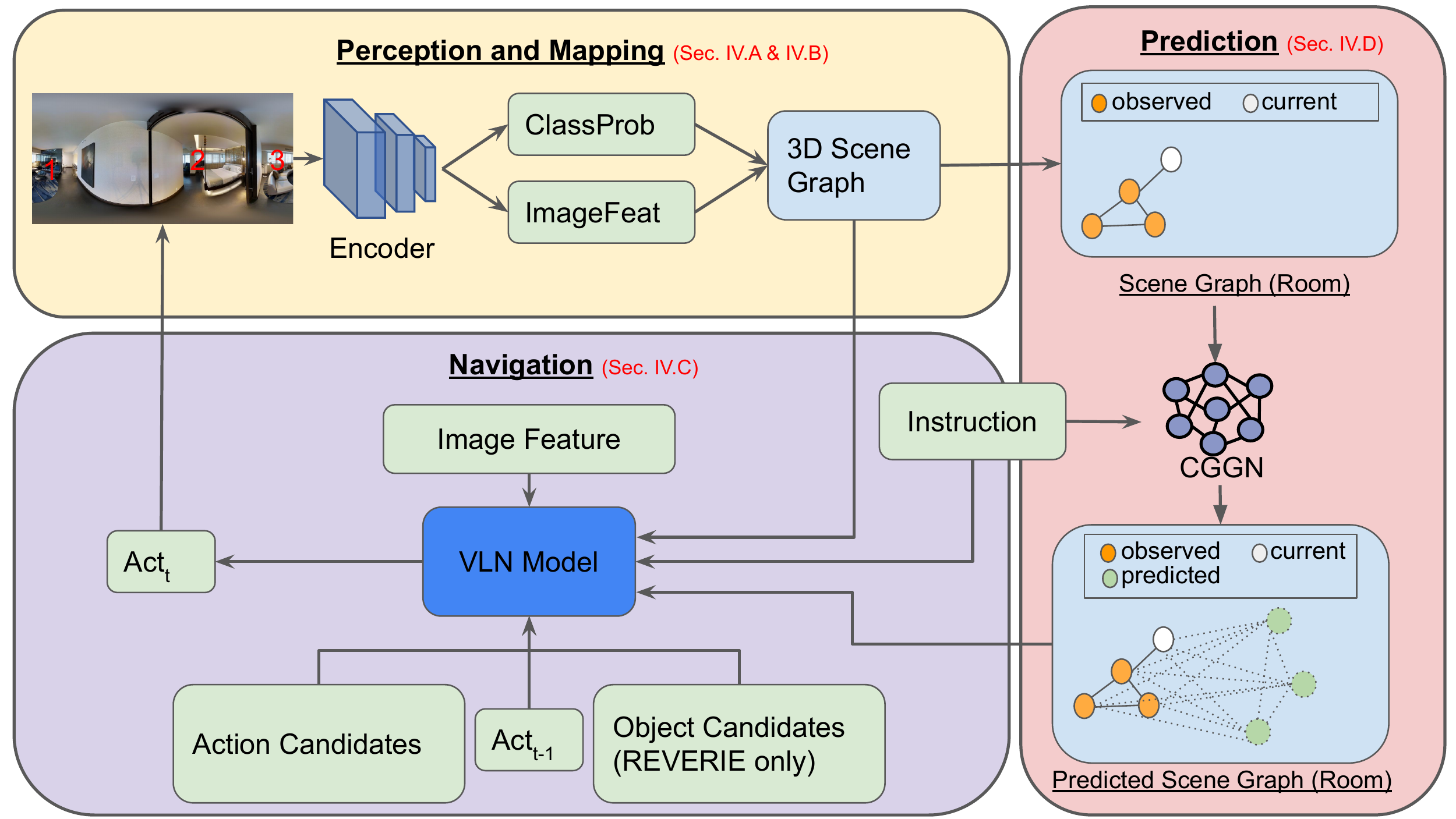}	
		\caption{
		    Schematic of the proposed navigation framework.
		    \textbf{Perception and Mapping: }
		    building a multi-level 3D scene graph with the use of predictions from perception models.
		    \textbf{Prediction: } 
		    predicting a probabilistic path leading to the goal.
		    \textbf{Navigation: } 
		    The relevant information is passed to the visual-language navigation (VLN) model for action prediction.
		}
		\label{fig:framework_overview}
\end{figure*}

Unlike much of the prior art \cite{mattersim,moghaddam2021optimistic,qi2020reverie,zhu2017target}, which mostly aims to solve this unseen navigation problem in an end-to-end learnt fashion -- from current percepts directly to local actions, 
our solution is explicitly built around three components: Perception and Mapping -- Prediction -- Navigation. This is illustrated in \figref{fig:framework_overview}, 
and in more detail in \secref{sec:method}.
The robot learns aspects of each of the different modules from a dataset of diverse indoor environments. 
The centerpiece of the framework is a probabilistic 3D Scene graph, which is incrementally constructed by the Perception and Mapping module, augmented by a conditional graph generation network (CGGN) in the Prediction stage, and finally used as a map prior to informing the Navigation module. 
The framework aims to make high-quality predictions about the building layout (conditioned on what has been seen so far and on past experience) to assist the navigation module in making better local choices and reducing search times.
%
we explicitly expose a topological representation of the world to the navigation module, in comparison with the end-to-end systems that can at best hope this knowledge is implicitly captured by their black-box network.
With the use of the topological map, we
(i) prevent unnecessary revisit to the explored regions;
(ii) predict subgoals to assist navigation;
(iii) create a system that is less inscrutable and more easily understood, both in success and failure.

\emph{\bf Contributions: }
In summary, we propose a method for goal-directed navigation in a previously unseen environment that can leverage \textit{familiarity} with similar environments obtained through topological graph prediction.
In creating this system we make the following contributions:
\begin{itemize}
	\setlength{\itemsep}{0pt}
	\setlength{\parskip}{0pt}
	\setlength{\parsep}{0pt}
    \item We propose a navigation framework incorporating the usage of perception and 3D scene graph, which leads to an explainable learning-based navigation system.
    \item We propose CGGN aiming to make sensible predictions about the likely unseen layout, 
    conditioned on the observed graph and instruction,
    to assist navigation.
    \item We construct a 3D Scene Graph incrementally as the robot explores a new environment. This graph memory is used to avoid repeated redundant exploration.
    \item We show through navigation tasks 
    on the Matterport3D dataset that these innovations lead to a higher  success rate of navigation in unseen environments. 
\end{itemize}
    
    \section{Related Work}
\label{sec:lit_review}
%
\emph{\bf Scene Recognition: }
The general pipeline of most scene recognition algorithms consists of three stages, feature extraction, transformation, and classification\cite{xie2020scene}.
Early algorithms mainly relied on hand-crafted descriptors like SIFT \cite{lowe2004distinctive}, SURF\cite{bay2006surf}.
In recent years, many large datasets \cite{lazebnik2006beyond, zhou2017places, zhou2014learning, xiao2010sun} for scene recognition become publicly available, in which Place365 \cite{zhou2017places} is the largest.
Meanwhile, the surge of deep learning methods in the last decade arouses researchers' interest in applying deep learning methods for solving scene recognition tasks.
\cite{yang2015multi, xie2015hybrid} are some of the early works that apply multi-scale CNN for scene recognition tasks. 
Most recently, \cite{Zhou21borm, Miao2021ots} propose to use object representation for indoor scene representation. 
We refer readers to the survey \cite{xie2020scene} for a detailed introduction.

\emph{\bf 3D Scene Graph: }
A scene graph is an abstract-level representation of a scene with constituent parts such as objects and regions.
Armeni \etal \cite{armeni_iccv19} propose using a 3D scene graph as a comprehensive semantic understanding of a scene.
The proposed 3D scene graph structure spans entire buildings, and a 4-layer structure is introduced for the 3D scene graph.
Starting with the highest ``Building" level, down to ``Rooms", ``Objects", and ``Cameras" levels,
these four layers represent semantics, 3D space, and cameras in the building.
The authors propose a geometric method to create a 3D scene graph automatically.
Similarly, \cite{Rosinol2020} presents a 3D dynamic scene graph that extends the notion to represent dynamic objects.
Different from \cite{armeni_iccv19,Rosinol2020}, who pre-build 3D scene graphs in explored environments, we build and predict the scene graph in unexplored environments in an online fashion based on observation.

\emph{\bf Graph Generation: }
Recently, there is increasing attention on using neural networks for building graph generative models \cite{liao2019gran,segler2018generating,gomez2018automatic,you2018graphrnn}.
Modeling graph generation problem with deep generative networks allows the model to learn structural information from the data and model graphs with complicated topology.
Recent auto-regressive frameworks show scalable and general performance. 
GraphRNN \cite{you2018graphrnn} is one of the pioneer auto-regressive frameworks that generates one entry or one column in a graph adjacency matrix at a time through an RNN. 
%
Most of the prior art focuses on the graph's edge generation problem by generating the graph adjacency matrix. 
The goal is to train a model representing the graph distribution in the training set \ie generate graphs similar to the training graphs.
However, our problem differs slightly from the typical graph generation problem in two aspects.
(1) we need to generate nodes with attributes \eg region class;
(2) the generation process is conditional: it depends on the observation graph and the destination instruction.
We introduce a conditional graph generation network that predicts node attributes simultaneously.

\emph{\bf Visual Navigation: }
Robot navigation in indoor environments is a long-standing problem. 
We refer readers to \cite{kostavelis2015semantic} in which a comprehensive overview of related classical work is provided.
The classical approach to robot navigation separates the tasks of mapping and tracking from path planning, while most recent deep learning methods are end-to-end solutions that train models to map visual input to robot actions.
\cite{zhu2017target,mirowski2016learning} propose end-to-end deep reinforcement learning methods that navigate to the goal location without explicitly building the map of the environment.
On the other hand,  \cite{savinov2018semi} introduces a memory module for navigation that stores visited locations. 
When short footage of a previously unseen environment is provided, the memory module builds a topological map to assist navigation. 

Some prior works explored using pre-computed navigation/topological/scene graphs to assist navigation.
They pre-compute a topological graph 
\cite{Chen2019, vln-pano2real, mccammon2021topological}
, an object scene graph \cite{yang2018visual,moghaddam2021optimistic}, and 3D scene graph \cite{ravichandran2021hierarchical} to assist navigation. 
The graph representations are embedded into an agent-centric feature space such that the agent learns a navigation policy conditioned on the graph prior.
\cite{wu2019bayesian} proposes a Bayesian Relational Memory for learning layout prior from training environments.
Similarly, \cite{beeching2020learning} predicts the shortest path from an estimated topological map. 
An agent is trained to navigate in an environment with the use of the predicted trajectory.
However, \cite{wu2019bayesian, beeching2020learning} assume that an explorative rollout is performed to construct a partial map in the first place. 
In contrast, we do not pre-build the graph but build and predict a multi-level topological map for the explored and unseen areas on-the-fly.
\cite{rincon2019time, Chaplot2020} shares same idea of building a topological map on-the-fly.
\cite{rincon2019time} 
builds a toposemantic map to facilitate exploring unexplored regions for reaching a given target location. 
\cite{Chaplot2020} builds a single-level topological map as the agent receives observations and predicts some unexplored nodes
directly connected to the built map. 
Instead of predicting unexplored nodes that are directly connected to the map, we forecast a trajectory that leads to the target goal.
With a similar motivation of forecasting the environment, Pathdreamer \cite{koh2021pathdreamer} forecasts full RGB and depth images for the future path.
\cite{narasimhan2020seeing,purushwalkam2020audio} predict top-down semantic floor plans and use the predicted maps for room navigation tasks. 
Although these methods forecast unseen regions, the semantic floor plan is a less efficient representation of building structure when compared to topological maps.

\emph{\bf Visual-And-Language Navigation: }
Recently Anderson \etal \cite{mattersim} have proposed the Visual-And-Language Navigation (VLN) task, which aims to ask a robot to navigate in an environment following human instructions. 
Specifically, \cite{mattersim} describes the R2R (Room-to-room) task in which the final goal (such as ``navigate to the kitchen'') is supplemented by intermediate instructions 
(e.g. ``exit the hall; turn right and walk to the end''). 
Solutions to this task include \cite{mattersim}'s baseline, as well as  \cite{fried2018speaker, wang2019reinforced, keLBHGLGCS19, hong2020recurrent}. 
Based on R2R, Qi \etal \cite{qi2020reverie} proposed the REVERIE task 
in which a robot is given a natural language instruction referring to a remote object and the robot must navigate to the correct location and identify the object.

Although these works provide an excellent environment for testing navigation, 
aspects of their tasks are slightly different from ours. 
In the R2R case, this is because of the availability of intermediate instructions from an entity (human) that is familiar with the scene layout. 
In our scenario only the high-level goal is specified: a restricted form of R2R which we call ``Area-Goal''. 
In the REVERIE case, there are no intermediate instructions, but the goals often require the agent to understand nuanced language (the referring expressions). 
So that we can make a direct comparison with the prior art, we adopt the REVERIE task as our main comparator, making use of a language model to convert the instruction into a latent vector that encodes room and object goals (to save space, the simpler Area-Goal task is explained and evaluated in the supplementary material).

\section{Problem Statement}
\label{sec:problem}
We consider the problem of instruction-directed exploration for visual navigation in an unseen environment: 
a robot is tasked with navigating to a goal destination, given an instruction that is specified at a much higher level of abstraction, 
\eg
``go to the kitchen/dining room (and get the spoon)".
We assume the robot has access to 
sensor data at each location, including visual data (color image) as the primary observation and position data (3D coordinates) \wrt the origin of exploration.

Matterport3D \cite{Matterport3D} is an RGB-D dataset containing 10,800 panoramic views captured from 90 building-scale scenes with a variety of sizes and complexity.
For training an agent to perform visual navigation, an environment simulator is usually required. 
Anderson \etal \cite{mattersim} introduce Matterport3D Simulator and R2R navigation task, which are specifically designed for navigation between panoramic viewpoints in Matterport3D Dataset.
Based on R2R, Qi \etal \cite{qi2020reverie} propose REVERIE dataset/task 
in which a robotic agent is given a natural language instruction referring to a remote object and the agent must navigate to the correct location and identify the object from the environment.
In this paper, our focus is on showing how we can learn and leverage similar environments to confer a degree of familiarity. To enable a direct comparison to similar systems that do not have access to this prior knowledge, we adopt the REVERIE setting, which is closest to ours. 
    
    %
\begin{figure*}[t!]
        \centering
\includegraphics[width=0.8\textwidth]{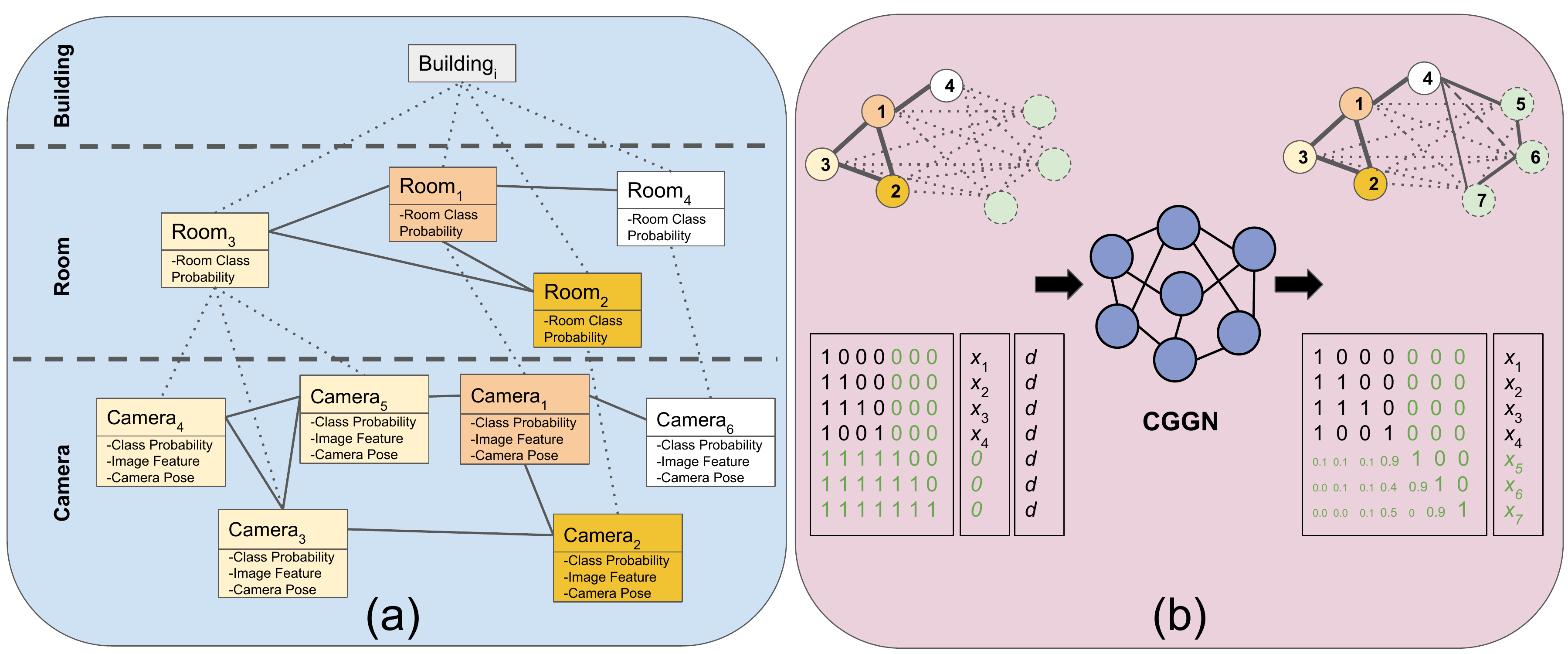}
	
		\caption{
		    (a) Multi-level 3D Scene Graph representing a topological map of the observed environment.
		    (b) The node and edge states of the input graph are initialized with the observed room-level graph and a destination vector, $d$. CGGN updates the node states and predicts a probabilistic 
		    trajectory graph. Green variables are associated to the predicted nodes.
		}
		\label{fig:3d_scene_graph_CGGN}
\end{figure*}

%
\section{Method}
\label{sec:method}
Scene understanding, navigation history, and subgoals can be useful information for navigation as they provide strong navigation guidance.
We introduce a framework (\figref{fig:framework_overview}) that utilizes these elements to assist navigation.
%
Our proposed framework has three stages, including \textit{Perception and Mapping}, \textit{Prediction}, and \textit{Navigation}.
In the Perception and Mapping stage, 
we aim to construct a multi-level scene graph representing the explored regions in a bottom-up manner as the agent explores the environment.
Next, the room-level graph and a user-specified instruction are passed to a conditional graph generation network (CGGN), which 
generates a new graph consisting of the observation part and a predicted trajectory graph leading to the destination.
This predicted graph is served as a \textit{graph prior} for navigation.
The relevant information is passed into a navigation module for informed action decision-making. 
Due to the space limit, we describe a higher-level idea of the proposed method in the main paper while more method details can be found in the supplementary material.

\subsection{Multi-level 3D Scene Graph}
The centerpiece of this framework is a 3D scene graph that is built and predicted as the agent explores the environment (\figref{fig:3d_scene_graph_CGGN}).
This 3D scene graph represents a topological map of the observed environment in three levels, Building -- Room -- Camera.
The \textit{building}-level includes the root node in the graph.
The \textit{rooms} of the building compose the second level.
Each room is represented with a unique node on this level.
The nodes in the lowest \textit{camera}-level graph includes camera locations and the attributes related to the observations (\eg room type and image features).
The connections between nodes within each level represent the traversability between nodes while the dahsed connections across different levels indicate the composition relationship.
%
\subsection{Perception and Mapping: Building 3D Scene Graph} \label{sec:method:perception}
We describe the operation of the perception module (yellow box in \figref{fig:framework_overview}), in particular the construction of the 3D Scene Graph, in this section.
During each episode, at each time-step, the agent receives a panoramic observation with a known 3D location.
We initialize a new node for the observation in the camera-level graph.
The node attributes include the camera location (in 3D relative to the starting point), along with the image features and the soft-max scores from a ResNet room classifier (the latter of which we treat as a probability distribution).
Furthermore, the agent tries to associate the camera node with a node in the room-level graph by comparing it against existing camera nodes.
To achieve this, we employ a binary MLP-classifier to predict the likelihood that two camera nodes are from the same room.
Once the camera node is localized in the room-level graph, we update the class probability of the room node by fusing their individual class probabilities, \ie by multiplying them and normalizing the final values.
Otherwise, a new room node will be initialized if the camera node is found to be in a new room.
The aim of localizing new observations in the scene graph is to prevent creating an infinite chain graph when the agent revisits previous nodes.

\subsection{Navigation} \label{sec:method:navigation}
In this section, we describe the operation and implementation of the navigation module (purple box in Figure \ref{fig:framework_overview}), whose purpose is to determine the next action for the agent. The key element to our framework is that both the current 3D Scene Graph and predicted graph are fed into the module; this notion is independent of the exact implementation of the module and can be adapted to a wide variety of navigation tasks.
Another example showcasing a different navigation task appears in the supplementary material.
However, we describe here the detail of a navigation module purposed for the REVERIE task  \cite{qi2020reverie} as described in \secref{sec:problem}.
It is based on a state-of-the-art \vlnbert \cite{hong2020recurrent} agent, which 
is trained to navigate between panoramic viewpoints to reach the correct location and identify the object from the environment given abstract instruction.
At each time-step, the agent is allowed to perform an action chosen from: 
(i) move to a nearby panoramic viewpoint candidate; 
or (ii) choose an object from a set of observed candidates at the current location.
If an object candidate is chosen, the agent is considered to be entering the ``Stop'' state with the chosen object as the target.


\emph{\bf Perception-informed exploration: }
The current percepts (image features and room-class probabilities) are explicitly evaluated by the navigation module to prevent stopping if the current location is unlikely to be the goal class.
Specifically, we do not force the robot to stay in a room when it is predicted to be the destination room type. It is because there can be multiple rooms with the same destination room type. 
All ``Stop'' actions are excluded if the goal room type is not in the top-$k$ predictions from the perception module. 
 Therefore, the robot remains a possibility to explore even if the current room type matches the destination room type.  

\emph{\bf Map-informed exploration: }
Exploration and navigation efficiency can be greatly reduced if an agent keeps revisiting explored regions.
Our 3D scene graph serves as a topological map/memory. 
We encourage the agent to visit unexplored regions by prioritising unexplored viewpoint candidates over visited ones in the action space. 
This priority order for actions prevents the agent from bouncing back and forth between locations, and in the worst-case, ``forces" a depth-first exploration of panoramas.
When there are no unexplored viewpoints in the candidates, the agent has to trace back to the closest location that has unexplored viewpoint candidates. 
In such a case, the robot could retraverse explored regions for exploring new regions. 
Arguably a ``softer'' method could be more effective, but our simple implementation does not detract from the main objective of informing search using predictions.  

\emph{\bf Subgoal-informed \vlnbert: }
Navigating from a random place to a destination in an unseen environment can be difficult 
while navigating via subgoals leading to the destination can be a relatively easier task.
Our Prediction module predicts a probabilistic trajectory leading to the destination,
which will be detailed in \secref{sec:method:prediction_graph}.
For now, we assume that we can provide the navigation module with the subgoal information extracted from the Prediction module.

Original \vlnbert  takes four sets of tokens as input:
previous state token $\textbf{s}_{t-1}$, 
language tokens $\textbf{X}$,
visual tokens for scene $\textbf{V}_t$
%
and visual tokens for objects $\textbf{O}_t$.
We add a token for the subgoal $\textbf{G}_{t}$, 
initialized with the room-type probability of the first node in the predicted trajectory.
Thus, a self-attention mechanism is employed to capture the textual-visual correspondences from the cross-modal tokens to predict the action probabilities $\textbf{p}_t^a$ and the object grounding probabilities $\textbf{p}_t^o$,
\begin{equation}
\begin{aligned}
    \textbf{s}_t, \textbf{p}_t^a, \textbf{p}_t^o
        &= 
        \text{\vlnbert} (\textbf{s}_{t-1}, \textbf{X}, \textbf{V}_t, \textbf{O}_t, \textbf{G}_t).
\end{aligned}
\end{equation}
%

We follow the training strategy used in \cite{hong2020recurrent}.
A mixture of reinforcement learning (RL) and imitation learning (IL) objectives is applied in the training. 
A2C \cite{pmlr-v48-mniha16} is used for RL.
The agent samples an action, $\textbf{a}_t^s$, according to $\textbf{p}_t^a$ and measures the advantage $\textbf{A}_t$ at each step.
For IL, the agent is trained with a cross-entropy loss by following the teacher actions, $\textbf{a}_t^*$, and object grounding, $\textbf{o}_t^*$, defined on the ground-truth trajectories.  
Overall, the following navigation loss function is minimised,
\begin{equation}
\begin{aligned}
    \mathcal{L}
        &= 
        - \sum_{t} 
        \textbf{a}_t^s \text{log} (\textbf{p}_t^a) \textbf{A}_t
        - \lambda \left(
        \sum_t
        \textbf{a}^*_t \text{log}(\textbf{p}_t^a)
        +  \sum_t
        \textbf{o}^*_t \text{log}(\textbf{p}_t^o)
        \right).
\end{aligned}
\end{equation}
%

\subsection{Prediction: Forecasting Subgoals} 
\label{sec:method:prediction_graph}
Humans have an impeccable ability to build an abstract-level map of the environment and use the experience to adapt to the new environment with the observed cues. 
Intuitively, 
when a person has observed a dining room, it is natural to guess that kitchen is nearby
(note, e.g., that \cite{narasimhan2020seeing} shows that architectural patterns can be learnt).
We formulate this problem as a \textit{graph generation} problem
and propose a conditional graph generation network (CGGN) that predicts unexplored environment conditioned on the observation graph and a destination instruction.
The room-level topological map can be employed naturally as the observation graph here. 
This aim for the prediction is to allow the agent to plan ahead through unobserved regions by ``speculating'' in an informed way what choices are likely to lead to the destination.  
We therefore formulate the graph generation task as generating a trajectory that leads to the destination
conditioned on the observations.

{\bf{Generation: }}
Our network takes two inputs: 
(i) destination instruction, represented by a one-hot vector of the destination class $d$;
(ii) the room-level graph observed to date, whose 
node attribute is the room-type probability, $x_{o}$,
and 
edges are represented by a lower triangular adjacency matrix $L_\mathbf{o}$ (the graphs are undirected).
The output of the network is a predicted trajectory graph that is connected to the observation graph.
The trajectory graph comprises nodes and edges as follows:
(i) node attribute is a predicted room-type probability, $x_\mathbf{t}$;
(ii) edges are one/few new row entries in addition to the observed adjacency matrix, where the new entries are denoted as $L_\mathbf{t}$.
The generation process is then formulated as
$L_\mathbf{t}, x_\mathbf{t}=\text{CGGN}(L_\mathbf{o}, x_\mathbf{o}, d)$.
%

\emph{Node initialization}:
At the generation step, we first initialize the \textit{i}-th node representation, $h^0_i$, of the observation graph by embedding adjacency matrix $L_i$, destination vector $d$, and node class probability $x_i$ via a linear mapping, $h^0_{i} = W[L_i, x_i, d] + b, \forall i \in \mathbf{o}$, where $\mathbf{o}$ indexes the nodes in the observation graph.
The embedding reduces the feature size when the graph size grows.
For the new nodes to be generated, we initialize their state with zeros.

\emph{Edge initialization}:
We initialize the edges involving new nodes to be fully connected with all other nodes.
We create an edge feature for each edge by denoting the involvement of the new nodes. See the supplementary material for more details.

\emph{Prediction and Supervision}:
We adopt the graph neural network (GNN) architecture proposed in \cite{liao2019gran} for message passing in the graph.
The GNN is applied to our initialized graph to recurrently update the node representation, $h^j_i$, in \textit{j}-th round of message passing.
The updated hidden states are then used to predict edge probability $p(L_\mathbf{t})$, and node class probability $p(x_\mathbf{t})$.
We obtain the final node representations $h_i^R$ for each node \textit{i} after $R$ rounds of message passing.
The probability of generating edges is modeled via a mixture of Bernoulli distributions \cite{liao2019gran}:
\begin{equation}
\begin{aligned}
    p(L_{\mathbf{t}} 
        ) &= 
        \sum_{i=1}^{K} \alpha_k 
        \prod_{i \in \mathbf{t}}
        \prod_{1\leq j\leq i} \theta_{k,i,j},
\end{aligned}
\end{equation}
where 
$\mathbf{t}$ indexes the nodes in the trajectory graph, \\
$\alpha_1, ..., \alpha_K = \text{Softmax}\left( 
                    \sum_{i \in \mathbf{t}, 1 \leq j \leq i} \text{MLP}_\alpha(h_i^R-h_j^R)
                            \right)$, \\
$\theta_{1,i,j}, ..., \theta_{K,i,j} = \text{Sigmoid}\left(
                            \text{MLP}_\theta(h_i^R-h_j^R)
                            \right)$, 
and $K$ is the number of mixture components.

A 3-layer MLP is employed to predict the node class probability via $p(x_i)=\text{Softmax}(\text{MLP}_x(h_i^R))$, $\forall i \in \mathbf{t}$,
which is supervised by the cross-entropy loss. 
Note that given the same observation graph and destination, there can be multiple trajectories leading to the destination node.
However, with a large amount of data, the network captures the distribution of the possible trajectories conditioned on observation and destination.
The output of this stage is a graph consisting of the observed regions and an unexplored ``imagined" trajectory that leads toward the destination.
We demonstrate the use of this predicted trajectory in \secref{sec:method:navigation} and another task in the supplementary material.

\section{Experiment and Results}
\label{sec:experiment}

\emph{\bf Dataset: }
We perform our experiments on Matterport3D dataset\cite{Matterport3D} that includes 90 building-scale scenes.
The scenes have been divided into three splits for training (61), validation(11), and testing(18).
In our experiments, we follow this standard split protocol for data preparation.
The detail 
of generation and the split 
of the dataset for each task is included in the supplementary material.

\subsection{Perception and Mapping}
%

\begin{table}[!t]
    \centering
\begin{minipage}[t]{0.5\linewidth}\centering
\label{table:exp:room_network_choice}
\caption{
    Room classifier study
}

\resizebox{1\columnwidth}{!}{
\begin{tabular}{ l |  c |   }
        \hline
        Network &
        Top-5 Accuracy
        \\
        \hline
        ResNet-18 & \textbf{62.7} \\
        ResNet-50 & 59.5 \\
        DenseNet-161 & 41.9
        \\
        \hline
    \end{tabular}
} \label{tab:room_classfier_study}

\end{minipage}\hfill%
\begin{minipage}[t]{0.48\linewidth}\centering
\caption{Localization study}
\label{table:exp:localization}

\resizebox{1\columnwidth}{!}{
 \begin{tabular}{ l |  c  }
        \hline
        Variants &
        Accuracy
        \\
        \hline
        no pose & 76.37
        \\
        no encoding & 77.38
        \\
        w/ encoding (L=5) & \textbf{78.15}
        \\
        \hline
\end{tabular}
}
\end{minipage}
\end{table}



%
%
During the training the Perception model (ResNet classifier), 
we find that there is dataset imbalanced issue for this task.
Class imbalance means that we make use of focal loss \cite{lin2017focal}, having found that this gives marked improvement to the ability to recognise infrequent room types compared to the commonly used cross-entropy loss.
To properly evaluate the performance of our Perception model, we measure
average Top-\textit{k} accuracy over classes instead of average accuracy over all samples.
The top-\textit{5} accuracy are $63.7\%$ and $66.3\%$ for CE-loss and focal loss, respectively.
We have experimented with some network variants (using cross-entropy loss), including ResNet-18, ResNet-50, and DenseNet-161. 
The comparison is presented in \cref{tab:room_classfier_study}. 
We found that the performance of ResNet-18 is the best for this task and it is the lightest one. We measure average top-k accuracy over classes.

For the Localization model (binary classifier), the network input includes image features and the camera pose of the camera nodes. 
We experiment with a few variants regarding the input data and present the result in \tabref{table:exp:localization}.
Additional camera pose helps in classifying panoramas from the same region since some regions can have similar visual features though they are in different locations (\eg toilets).
We also experiment with different positional encoding schemes \cite{vaswani2017attention} and found that encoding with Fourier transform (length=5) gives the best result.

\subsection{Prediction}
%
%
%
%
\begin{table} [t]  
    \begin{center}
    \resizebox{0.8\columnwidth}{!}{%
    \begin{tabular}{ l |  c  c | c c }
        \hline
        &
        \multicolumn{2}{c|}{\textbf{Node}} & 
        \multicolumn{2}{c}{\textbf{Edge}} \\
        &
        Top-1 & 
        Top-5 &
        Accuracy &
        Recall \\
        \hline
        Validation & 
        58.8 
        & 83.6 &
        97.0 & 86.4
        \\
        Test &  
        57.2 
        & 82.0  &
        96.0 & 82.5
        \\
        \hline
        \end{tabular}
    }
    \end{center}
    \vspace{-1em}
    \caption{Graph prediction result. 
    }
    \label{table:exp:graph_prediction}
\end{table}
%

%
Evaluating the quality of the predicted graphs is difficult for several reasons:
(1) 
finding a proper metric to evaluate the graphs.
Though graph similarity measures like graph edit distance \cite{ged1983} can be applied, the evaluation is biased by the edges due to the imbalanced ratio between edges and nodes;
(2)
comparison between a predicted soft graph and a GT hard graph;
(3)
multiple possible routes that lead to the destination. 
Considering these difficulties, we evaluate edges and nodes separately and report the result in \tabref{table:exp:graph_prediction}.

We evaluate the node classification result by computing the top-k accuracy.
Note that we compare our predicted trajectory with a maximum length of 5 nodes against all possible ground-truths and report the best result.
In order to determine the number of nodes to be predicted, we have experienced different values (\ie 3, 5, 10, 15, 20) and found that 5 nodes have the best result. 
The comparison is presented in \tabref{table:exp:trajectory_length}.
\begin{table}
    \begin{center}
    \resizebox{1.\columnwidth}{!}
    {%
    \begin{tabular}{ l |  c | c |  }
        \hline
        Number of nodes &
        Node (Top-1 Accuracy) &
        Edge (Recall)
        \\
        \hline
        3 & 42.5 & 86.3
        \\
        5 & \textbf{58.8} & \textbf{86.4}
        \\
        10 & 48.9 & 72.5
        \\
        15 & 46.9 & 72.9
        \\
        20 & 51.4 & 74.6
        \\
        \hline
        \end{tabular}
    }
    \end{center}
    \caption{
    Study for the choice of the number of generated node
    }
    \label{table:exp:trajectory_length}
\end{table}
For edges, we sample 100 times from the predicted edge probability distribution and use the mode of samples as the final prediction.
Thus we compare this final prediction with the GT graph selected from the node evaluation.
To better reflect the performance, we also report the recall evaluation due to the sparsity of the adjacency matrix.
We provide more evaluation and ablation studies in the supplementary material.
\subsection{Navigation}
%
\begin{figure*}[t!]
        \centering
		\subfloat{\includegraphics[width=0.48\textwidth]{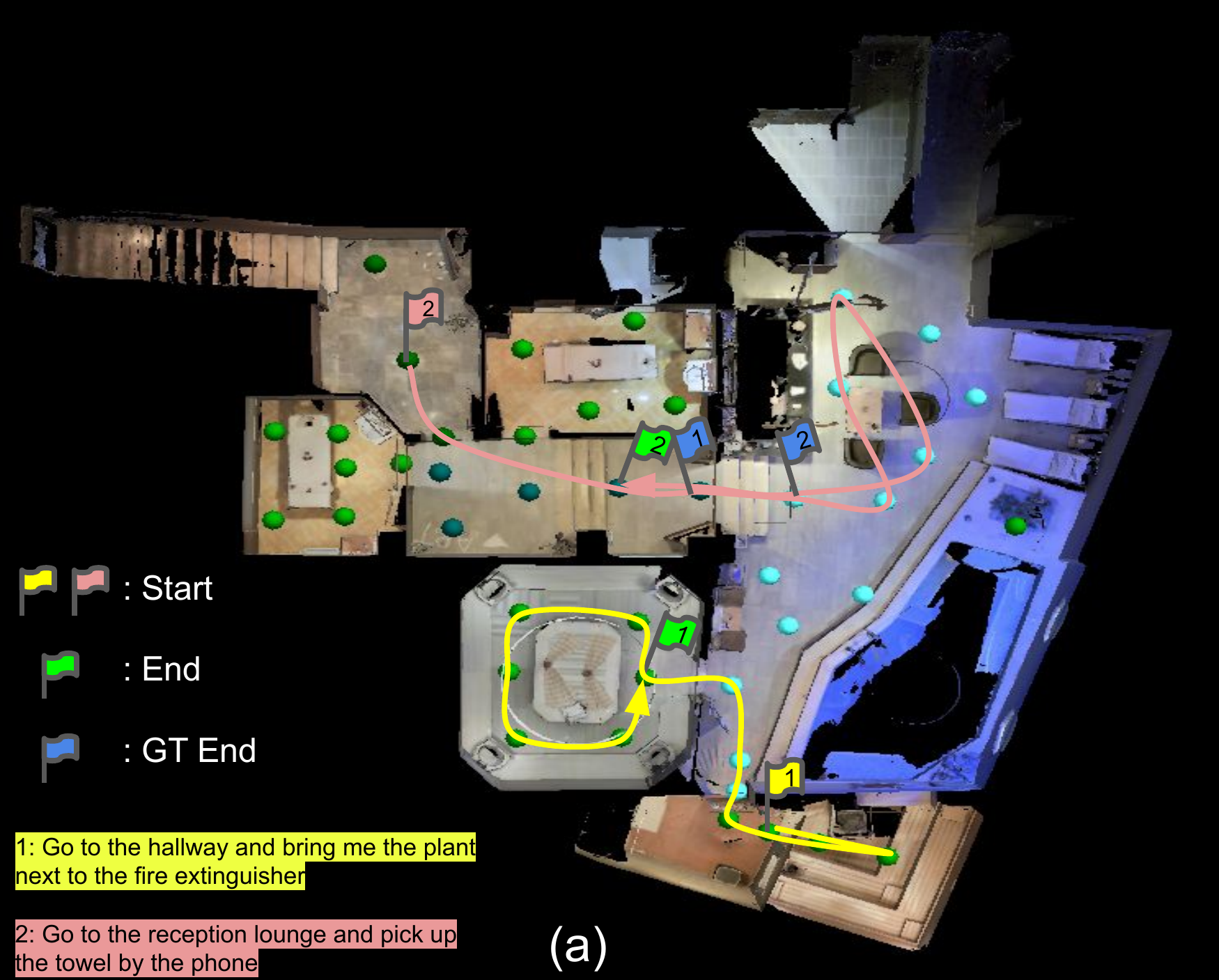}	\label{fig:nav_eg_baseline1}}
		\subfloat{\includegraphics[width=0.48\textwidth]{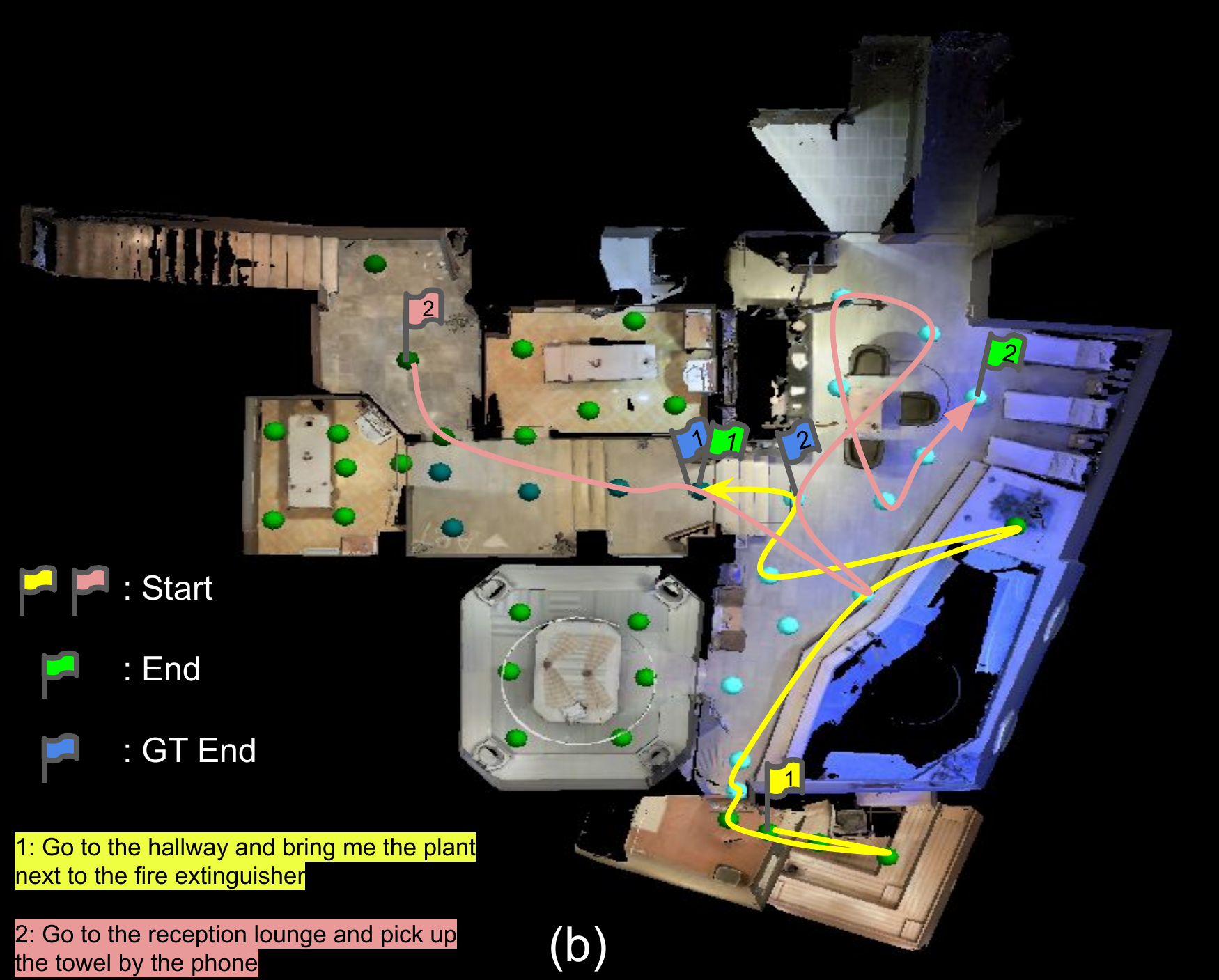}	
		\label{fig:nav_eg_ours1}}
	
		\caption{
		    Two instructed navigation examples are shown above, whose flags are labeled accordingly.
		    The colored circles represent panoramic viewpoints in different regions.
		    (a) Baseline
		    (b) Ours ([$\textbf{G}$, $\textbf{P}$, $\textbf{M}$]-informed).
		    Note that though the final destination differs from GT in trajectory-2, our agent navigates to the correct region and the target object is visible at the location.
		}
		\label{fig:nav_eg}
\end{figure*}

\emph{\bf Navigation metrics: }
REVERIE \cite{qi2020reverie} defines two sets of metrics for evaluating navigation and remote referring expression. 
Success Rate (SR) is defined as the ratio of stopping at a viewpoint where the target object is visible.
SPL considers the corresponding Success weighted by the normalized inverse of the Path Length. 
Oracle Success Rate (OSR) is the ratio of having a viewpoint along the trajectory where the target object is visible.
For object grounding evaluation, Remote Grounding Success Rate (RGS) is the ratio of grounding to the correct objects when stopped, and
RGSPL, which weights RGS by the trajectory length.

\subsubsection{Ablation Studies}
We use \vlnbert 
as Baseline and perform ablation studies against it. 
Our evaluation will focus on the Navigation metrics (SR and SPL) in unseen environments as we care more if the agent reaches the correct place in unseen scenes. 
The study result is summarized in \tabref{table:exp:ablation}.
A qualitative comparison is presented in \figref{fig:nav_eg}.
%

%
    %
\begin{table} [th!]  
    \begin{center}
    \resizebox{0.3\textwidth}{!}{%
    \begin{tabular}{ | l |  c  c | c c | } 
        \hline
        \multirow{2}*{\textbf{Experiments}} &
        \multicolumn{2}{c|}{\textbf{Val Unseen}} \\
        & SR & SPL \\
        \hline
        \hline
        Baseline 
        & 27.4 & 23.1 \\
        
        Baseline, 50steps
        & 27.9 & 23.2 \\
        
        [$\textbf{G}$]
        & 29.8 & 24.2 \\
        
        [$\textbf{G}$, $\textbf{P}$]  
        & 30.6 & 24.4  \\
        
        [$\textbf{G}$, $\textbf{M}$] 
        & 29.7 & 25.3 \\
        
        [$\textbf{G}$, $\textbf{P}$, $\textbf{M}$] 
        & 31.8 & 26.1  \\
        
        [$\textbf{G}$, $\textbf{P}$, $\textbf{M}$] , 50steps
        & 33.3 & 26.5 \\
        
        \hline
        \hline
        \multicolumn{3}{|c|}{Ground-Truth information involved} \\
        \hline

        [$\textbf{G}^*$] 
        & 31.0 & 25.5 \\
        
        [$\textbf{G}$, $\textbf{P}^*$]  
        & 33.1 & 26.1 \\
        
        
        [$\textbf{G}$, $\textbf{P}^*$] , 50steps
        & 34.5 & 26.3 \\
        
        
        [$\textbf{G}$, $\textbf{P}^*$, $\textbf{M}$]  
        & 37.9 & 29.4 \\
        
        [$\textbf{G}$, $\textbf{P}^*$, $\textbf{M}$] , 50steps
        & 45.1 & 30.1 \\
        

        
        \hline
    \end{tabular}
    }
    \end{center}
    \caption{
        Ablation studies. 
        $\textbf{G}^*$/$\textbf{G}$: GT/predicted subgoal;
        $\textbf{P}^*$/$\textbf{P}$: GT/predicted Perception; 
        $\textbf{M}$: Map-informed exploration;
        50steps: maximum 50 steps.
    }
    \label{table:exp:ablation}
\end{table}

%
\emph{\bf Subgoal-informed exploration: }
([$\textbf{G}^*$] \vs Baseline): 
We first validate that subgoal information is important for the navigation task by passing the GT subgoal token (one-hot vector representing the class of the next subgoal region) to \vlnbert
([$\textbf{G}^*$] \vs [$\textbf{G}]$]): Although there is still a performance gap caused by the accumulated errors in different stages from the framework, [$\textbf{G}$] still shows improvement over Baseline.
%

\emph{\bf Perception-informed exploration: }
We explicitly exclude ``Stop" action from the action space if the goal is not in the top-10 Perception predictions. 
We validate this mechanism is useful by comparing [$\textbf{G}$] against [$\textbf{G}$, $\textbf{P}$/$\textbf{P}^*$].

\emph{\bf Map-informed exploration: }
The agent stores the explored map in the form of a scene graph.
This graph allows us to enforce the agent prioritise visiting unexplored regions. 
([$\textbf{G}$, \textbf{M}] \vs [$\textbf{G}$]), 
and 
([$\textbf{G}$, $\textbf{P}$/$\textbf{P}^*$, \textbf{M}] \vs [$\textbf{G}$, $\textbf{P}$/$\textbf{P}^*$]) show that this map-informed exploration is helpful in navigation consistently.
We notice that VLN tasks usually set a step number (15) as the maximum steps that the agent can perform during each episode.
(Baseline \vs Baseline, 50steps):
Further increasing this limit does not help much.
However, as we build a map on the fly and encourage the agent to move to unexplored regions,
our agent has a higher chance to reach the correct destination, which is reflected by the high OSR in \tabref{table:exp:reverie}.
To further validate this argument, we compare 
([$\textbf{G}$, $\textbf{P}^*$, $\textbf{M}$] \vs [$\textbf{G}$, $\textbf{P}^*$, $\textbf{M}$, 50steps])
and 
([$\textbf{G}$, $\textbf{P}^*$] \vs [$\textbf{G}$, $\textbf{P}^*$, 50steps]).
The result suggests that the performance of more steps is greatly improved in the existence of the map while slightly improved without the map.

\emph{\bf Influence of Perception: }
%
%
([$\textbf{G}$, $\textbf{P}^*$] \vs [$\textbf{G}$, $\textbf{P}$]) and 
([$\textbf{G}$, $\textbf{P}^*$, $\textbf{M}$] \vs [$\textbf{G}$, $\textbf{P}$, $\textbf{M}$]): 
The performance gap in these comparisons suggests an interesting result that 
the navigation can be improved significantly given a better Perception model.
%
We find that 52\% of the failure cases of the Baseline model end up at a location which is not the room type of the goal.
This shows the importance of a Perception module that can guide the exploration/stop.
The wrong-room-rate is reduced to 42\% for $[\textbf{G}, \textbf{P}, \textbf{M}, \text{50steps}]$ model and 21\% for $[\textbf{G}, \textbf{P}^*, \textbf{M}, \text{50steps}]$ when the Perception is perfect.
%
%
\begin{table*} [th!]  
    \begin{center}
    \resizebox{1\textwidth}{!}{%
    \begin{tabular}{ l |  c  c c | c c  | c c c | c c  | c c c | c c  } 
        \hline
        \multirow{3}*{Methods} &  
        \multicolumn{5}{c|}{\textbf{Val Seen}}  & 
        \multicolumn{5}{c|}{\textbf{Val Unseen}}  & 
        \multicolumn{5}{c}{\textbf{Test Unseen}}  \\
        \cline{2-16}
        
        &
        \multicolumn{3}{c|}{Navigation} & 
        \multirow{2}*{RGS $\uparrow$} &  
        \multirow{2}*{RGSPL $\uparrow$} &  
        \multicolumn{3}{c|}{Navigation} & 
        \multirow{2}*{RGS $\uparrow$} &  
        \multirow{2}*{RGSPL $\uparrow$} &  
        \multicolumn{3}{c|}{Navigation} & 
        \multirow{2}*{RGS $\uparrow$} &  
        \multirow{2}*{RGSPL $\uparrow$}  \\
        \cline{2-4}
        \cline{7-9}
        \cline{12-14}
        
        &
        SR $\uparrow$ & OSR $\uparrow$ & SPL $\uparrow$ & & & 
        SR $\uparrow$ & OSR $\uparrow$ & SPL $\uparrow$ & & & 
        SR $\uparrow$ & OSR $\uparrow$ & SPL $\uparrow$ & & 
        \\
        \hline
        \hline
        
        RANDOM 
        & 2.74 & 8.92 & 1.91  & 1.97 & 1.31 
        & 1.76 & 11.93 & 1.01 & 0.96 & 0.56 
        & 2.30 & 8.88 & 1.44  & 1.18 & 0.78 \\
        
        Human
        & – & – & –  & – & – 
        & – & – & –  & – & – 
        & 81.51 & 86.83 & 53.66  & 77.84 & 51.44 
        \\
        \hline
        
        Seq2Seq-SF \cite{mattersim}
        & 29.59 & 35.70 & 24.01  & 18.97 & 14.96 
        & 4.20 & 8.07 & 2.84  & 2.16 & 1.63 
        & 3.99 & 6.88 & 3.09  & 2.00 & 1.58 
        \\
        
        
        SMNA \cite{ma2019self}
        & 41.25 & 43.29 & 39.61  & 30.07 & 28.98 
        & 8.15 & 11.28 & 6.44  & 4.54 & 3.61 
        & 5.80 & 8.39 & 4.53  & 3.10 & 2.39 
        \\
        
        
        FAST-MATTN \cite{qi2020reverie}
        & 50.53 & \textbf{55.17} & 45.50  & 31.97 & 29.66
        & 14.40 & 28.20 & 7.19  & 7.84 & 4.67
        & 19.88 & 30.63 & 11.61  & 11.28 & 6.08
        \\
        
        \vlnbert \cite{hong2020recurrent}
        & \textbf{51.79} & 53.90 & \textbf{47.96}  & \textbf{38.23} & \textbf{35.61} 
        & \underline{30.67} & \underline{35.02} & \underline{24.90}  & \underline{18.77} & \underline{15.27 }
        & \underline{29.61} & \underline{32.91} & \textbf{23.99}  & \underline{16.50} & \underline{13.51 }
        \\
        
        \hline
        Baseline
        & 49.54 & 50.67 & 45.12 & 36.56 & 33.40 
        & 27.35 & 30.56 & 23.07  & 17.18 & 14.51 
        & 28.40 & 30.61 & 22.97  & 15.03 & 12.20 
        \\

        \textbf{Ours}
        & \underline{51.45} & \underline{53.99} & \underline{46.12}  & \underline{36.91} & \underline{33.30}
        & \textbf{33.31} & \textbf{42.35} & \textbf{26.50}  & \textbf{19.82} & \textbf{15.54} 
        & \textbf{32.60} & \textbf{42.50} & \underline{23.52}  & \textbf{19.20} & \textbf{14.04 } 
        \\
        
        \textbf{Ours}*
        & 59.35 & 61.19 & 50.00 & 41.14 & 35.66 
        & 45.13 & 53.11 & 30.06  & 27.15 & 17.90
        & 40.07 & 55.10 & 24.64  & 23.41 & 14.68 
        \\

        \hline
    \end{tabular}
    }
    \end{center}
    \caption{
        REVERIE result comparison.
        \textbf{Ours} is our best model while \textbf{Ours}* assumes a perfect Perception module, which shows the potential of our proposed framework.
    }
    \label{table:exp:reverie}
    \vspace{-10pt}
\end{table*}

\pagebreak
\emph{\bf Comparison with SoTA: }
We present the comparison of our result against prior arts in \tabref{table:exp:reverie}.
Our model shows comparable result with \vlnbert in seen environments and outperform prior arts in unseen environments.
We want to emphasize that our framework has a potential to achieve better result if we can improve the Perception model (\textbf{Ours}*), which can be achieved with various ways such as a better backbone network and more data.

\section{Discussion}
\label{sec:discussion}
We have proposed an instruction-oriented navigation framework with a mixture of learning-based models and rule-based mechanisms.
The topological graph and informed trajectory predictions conditioned over this graph are the core novel contributions of our work. We have integrated these ideas into a framework for navigation, whose results demonstrate 
measurable benefit for robot navigation in unexplored environments.
Nevertheless, there are still some limitations in our proposed method.
(1) While we have demonstrated the method in the VLN Matterport 3D Simulator \cite{mattersim}, and this has forced on us assumptions about how the agent moves (it jumps between panoramic views instead of continuous motion), this further assumes that robot agent understands the traversability in the space.
However, \cite{vln-pano2real, Hong_2022_CVPR} have shown promising sim-to-real capability of instruction-goal agent running in continuous space when trained in discrete simulation.
We aim to adopt similar technique in \cite{vln-pano2real, Hong_2022_CVPR} for real-world robot employment in the future.
(2) Though our Prediction module provides a future subgoal to assist navigation, a limitation is that the agent cannot disambiguate between two rooms of the same class since the subgoal token does not provide other information (\eg relative location) to discover the difference between these rooms.
(3) Our proposed approach assumes that the robotic agents are capable of localizing themselves in the environment via well-developed localization methods, e.g. state-of-the-art SLAM/visual odometry methods
\cite{mur2017orb2, campos2021orb3, zhan2020visual}.


    


\section{Overview} \label{supp:sec:overview}
In this supplementary material, we provide more information on this work in the following aspects:
(1) more method details, including dataset, networks, and implementation details;
(2) more experiments and results;
(3) application of our proposed framework in Area-Goal navigation task;
(4) a video demonstration.

%
\begin{figure*}[t!]
        \centering
		\includegraphics[width=1\textwidth]{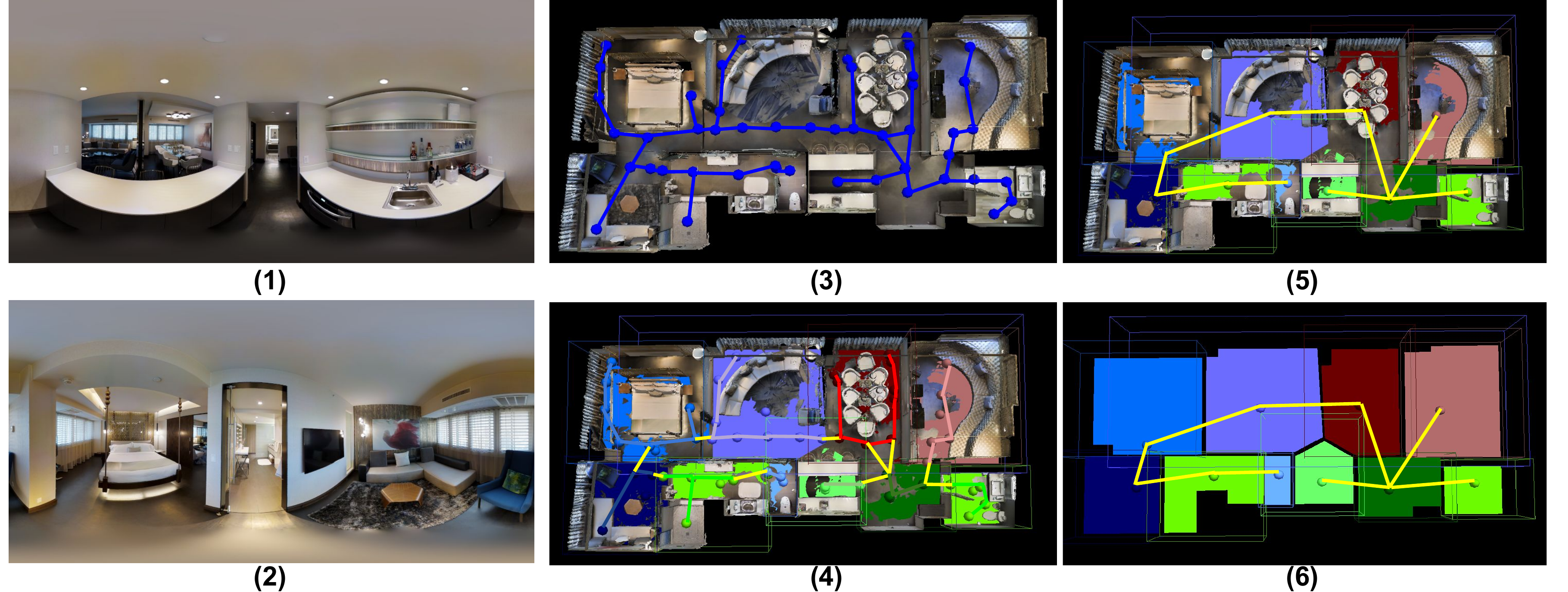}
		\caption{
		    Matterport3D dataset examples. 
		    \textbf{(1)(2)}:
		    Panorama examples.
		    \textbf{(3)}: 
		    Panorama connectivity graph \cite{mattersim}.
		    \textbf{(4)}: 
		    Intra/Inter-room panorama connectivity graph. Yellow edges represent the connectivity between panoramas captured at different rooms. It is further processed to generate the room connectivity graph.
		    \textbf{(5)(6)}:
		    Room connectivity graph. Each node represents a room, and the yellow edges represent the traversibility between rooms. Overlaying the graph on the full map and 2D room segmentation map are visualized.
		}
		\label{supp:fig:mp3d}
\end{figure*}
\section{Dataset} \label{supp:sec:dataset}
Matterport3D dataset is a large, diverse RGB-D dataset containing 10,800 panoramic views captured from 90 building-scale scenes with a variety of sizes and complexity.
The dataset comprises rich geometric and semantic information (\figref{supp:fig:mp3d}), in which we are interested in the region semantic annotation data in this work.
\textit{30} scene types (\eg bedroom, office \etc) are annotated for each region.
The scenes have been divided into three splits for training (61), validation(11), and testing(18).
We follow this standard data split protocol for data preparation in all our experiments.
We process the data for the model training in the sub-tasks as described in the following sections.

%
\begin{figure*}[t!]
        \centering
\includegraphics[width=0.7\textwidth]{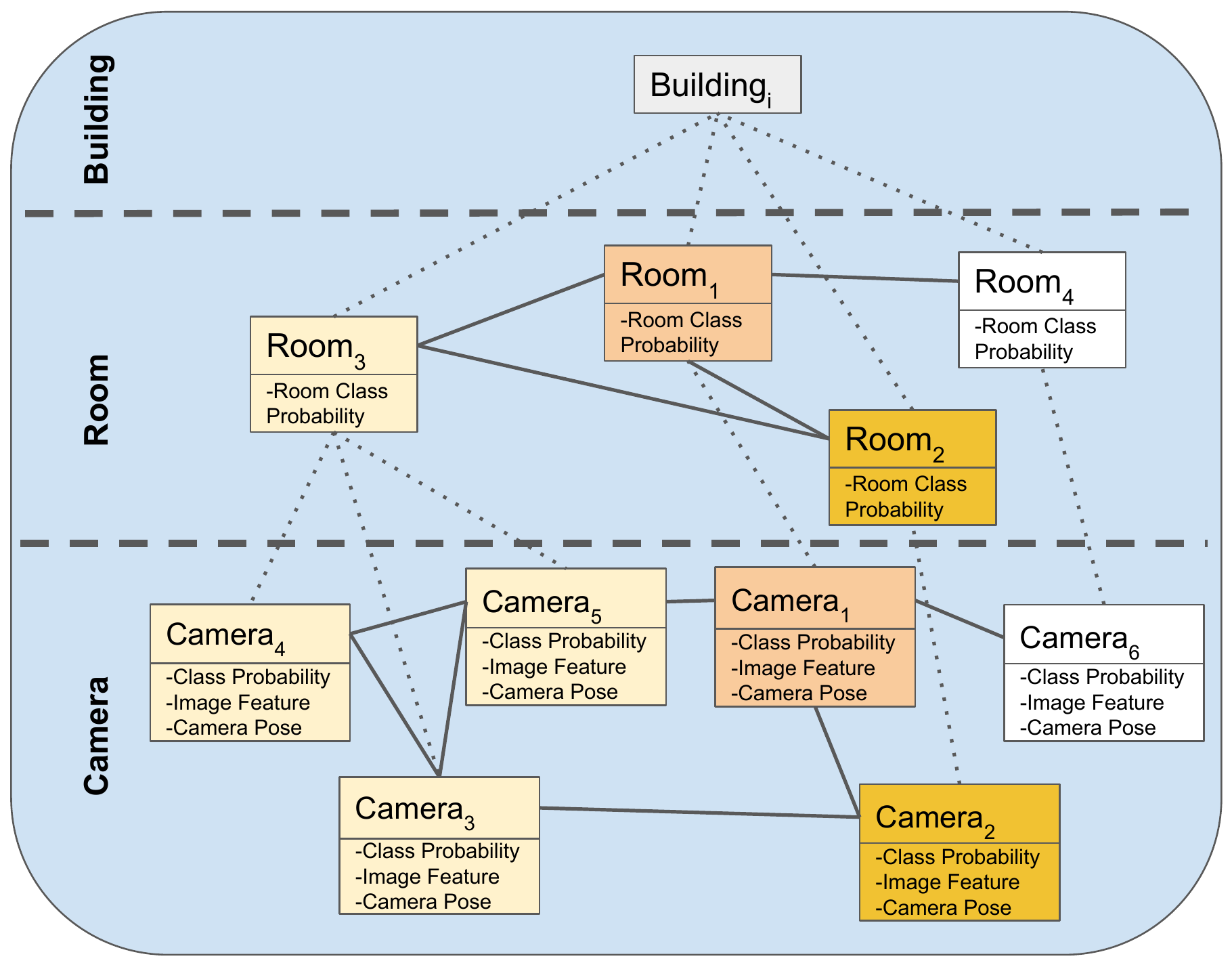}
	
		\caption{
		    (a) Multi-level 3D Scene Graph representing a topological map of the observed environment.
		}
		\label{supp:fig:3d_scene_graph}
\end{figure*}
\section{More Detail of \textit{Perception and Mapping}} \label{supp:sec:perception_mapping}
In this stage, we aim to construct a multi-level scene graph (\figref{supp:fig:3d_scene_graph}) as the robotic agent explores the environment.
Broadly, this consists of three separate steps:
(1) recognize the room type of the current location from visual observations (\ie panoramas);
(2) localize the panorama in a node in the room-level scene graph,
(3) update the region class probability in the corresponding node.
We model (1) and (2) as a multi-class classification and a binary classification problem, respectively.

Our perception model works with panoramas, which capture richer visual information in the local regions.
We initialize a new node for the observation in the camera-level graph.
A ResNet-18 network \cite{he2015resnet} is employed as the backbone of our perception model, which takes panorama as input and predicts room class probability.
We store the probability vector as one of the node attributes in the new camera node.
Meanwhile, the panorama feature extracted from the second-last layer of the perception model is stored as another node attribute.

In order to localize the new camera node in the room-level graph, we 
we pair the new camera node with existing camera nodes and predict the likelihood that the two nodes are from the same room via an MLP classifier.
If the pair with the highest likelihood has a score higher than 0.5, we consider they are from the same region. 
Otherwise, we create a new region node for the camera node.

Once the camera node is localized in the room-level graph, we update the class probability of the room node by fusing their class probabilities, \ie by multiplying them and normalizing the final values.
Otherwise, a new room node will be initialized if the camera node is found to be in a new room.

\subsection{Dataset}
We first generate panoramas from the raw skybox images.
For training the Perception model (room-type classifier), [7480, 933, 1945] panoramas are included in the training, validation, and test set, respectively.

For training the Localization model (binary classifier), we form positive and negative pairs by pairing panoramas from the same rooms and different rooms. 
Note that we have created equal numbers of positive and negative pairs to avoid an imbalance issue.
[54,294, 3,877, 15,103] panorama pairs are included in the training, validation, and test set, respectively.

\subsection{Network and Training details}
{\bf Room classifier: }
%
We adopt a ResNet-18 \cite{he2015resnet} as our backbone for the Perception model.
Specifically, we use the ResNet-18 model pre-trained in Place365 \cite{zhou2017places} as our model initialization.
Our best model is trained for 11 epochs using Adam, with a batch size of 100, an input resolution of $224 \times 224$, and a learning rate of $10^{-4}$.
Focal loss is used as the objective function and 
the gamma for the focal loss is $0.5$.

{\bf Binary classifier: }
\begin{figure}[t]
        \centering
		\includegraphics[width=0.8\columnwidth]{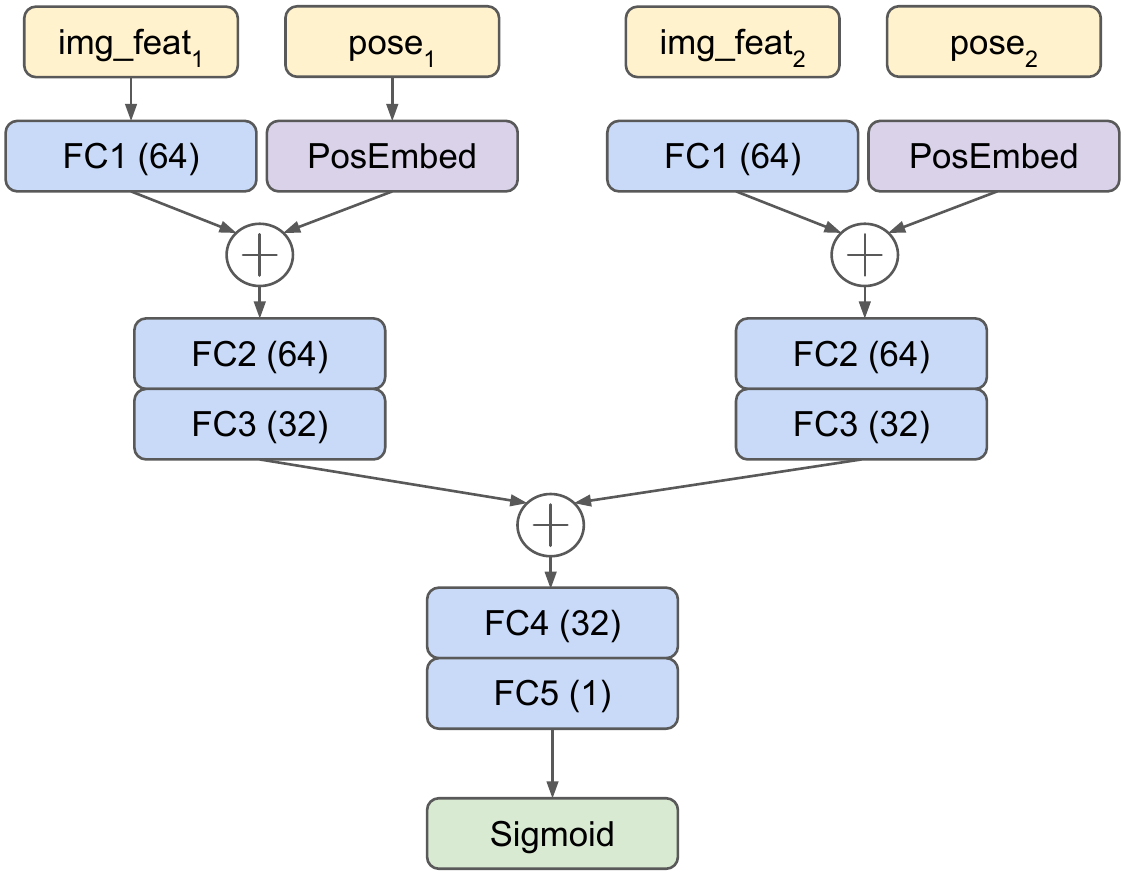}
		\caption{
		    Panorama Similarity Network. 
		    ReLU activation layers are applied for fully connected layers (FC1-4).
		}
		\label{supp:fig:pano_sim_net}
\end{figure}

For the binary classifier, we construct a multilayer perceptron that takes the camera pose and image feature from each panoramic view in the pair as input and predicts the likelihood that they are from the same region.
We further encode the camera pose using Fourier transform similar to the positional encoding technique in \cite{vaswani2017attention}.
The network architecture is shown in \figref{supp:fig:pano_sim_net}.
Our best model is trained for 11 epochs using Adam, with a batch size of 100, an input resolution of $224 \times 224$, and a learning rate of $10^{-4}$.
Though more advanced loss (like triplet loss variants) can be applied for this problem and possibly improve the result, we use the simplest binary cross-entropy loss, and it has already resulted in a model with 78\% accuracy.

\subsection{Experiment and Results}
\begin{figure}[ht]
        \centering
		\includegraphics[width=1\columnwidth]{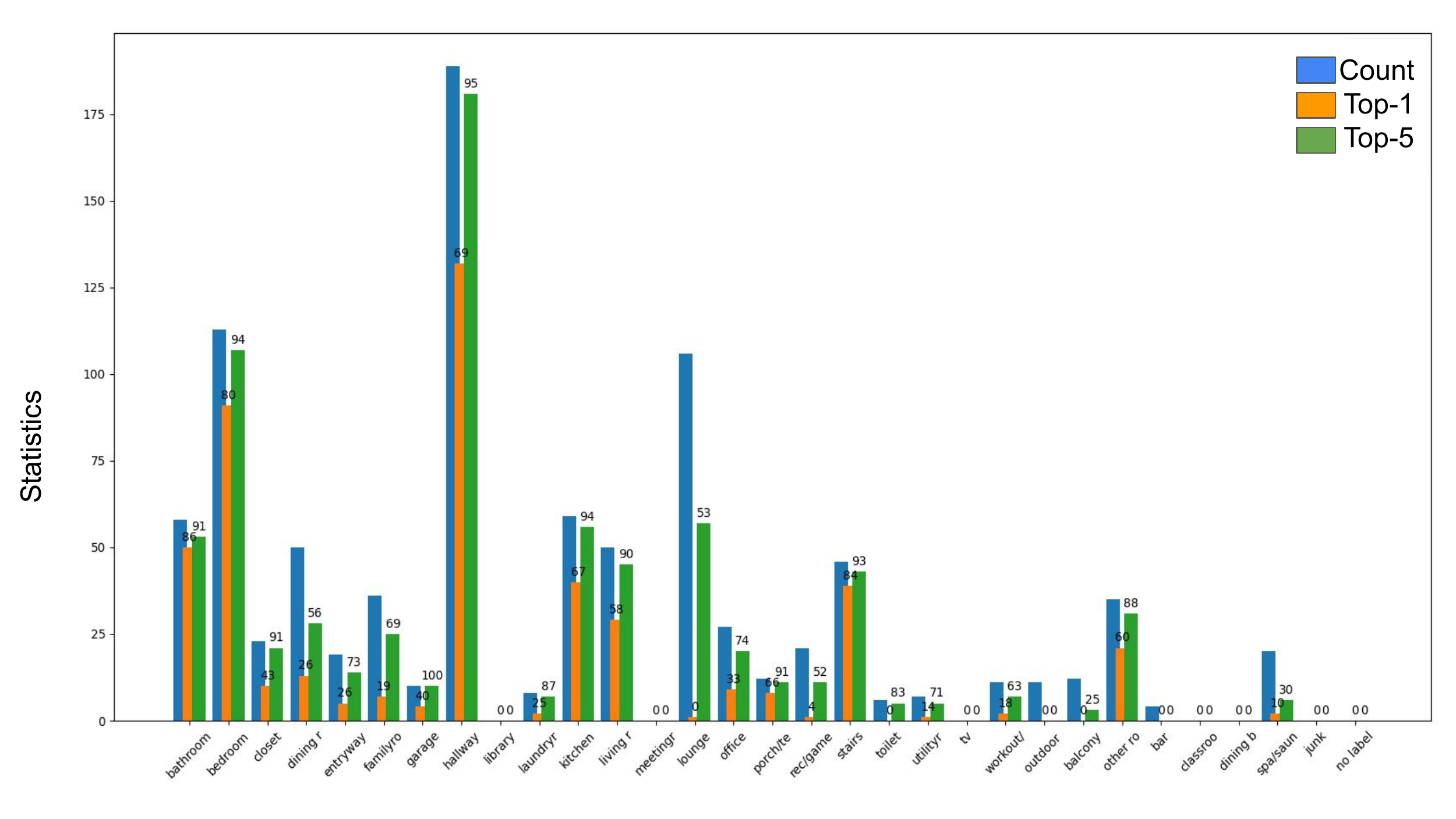}
		\caption{
		    Scene recognition result analysis. Best viewed with zoom-in. We count the appearance frequency of each class in the test set (blue bars) and their corresponding top-1/5 accuracy (orange and green bars). The accuracy is shown above the bars.
		}
		\label{supp:fig:sr_stat}
\end{figure}

\begin{figure}[th]
        \centering
		\includegraphics[width=1\columnwidth]{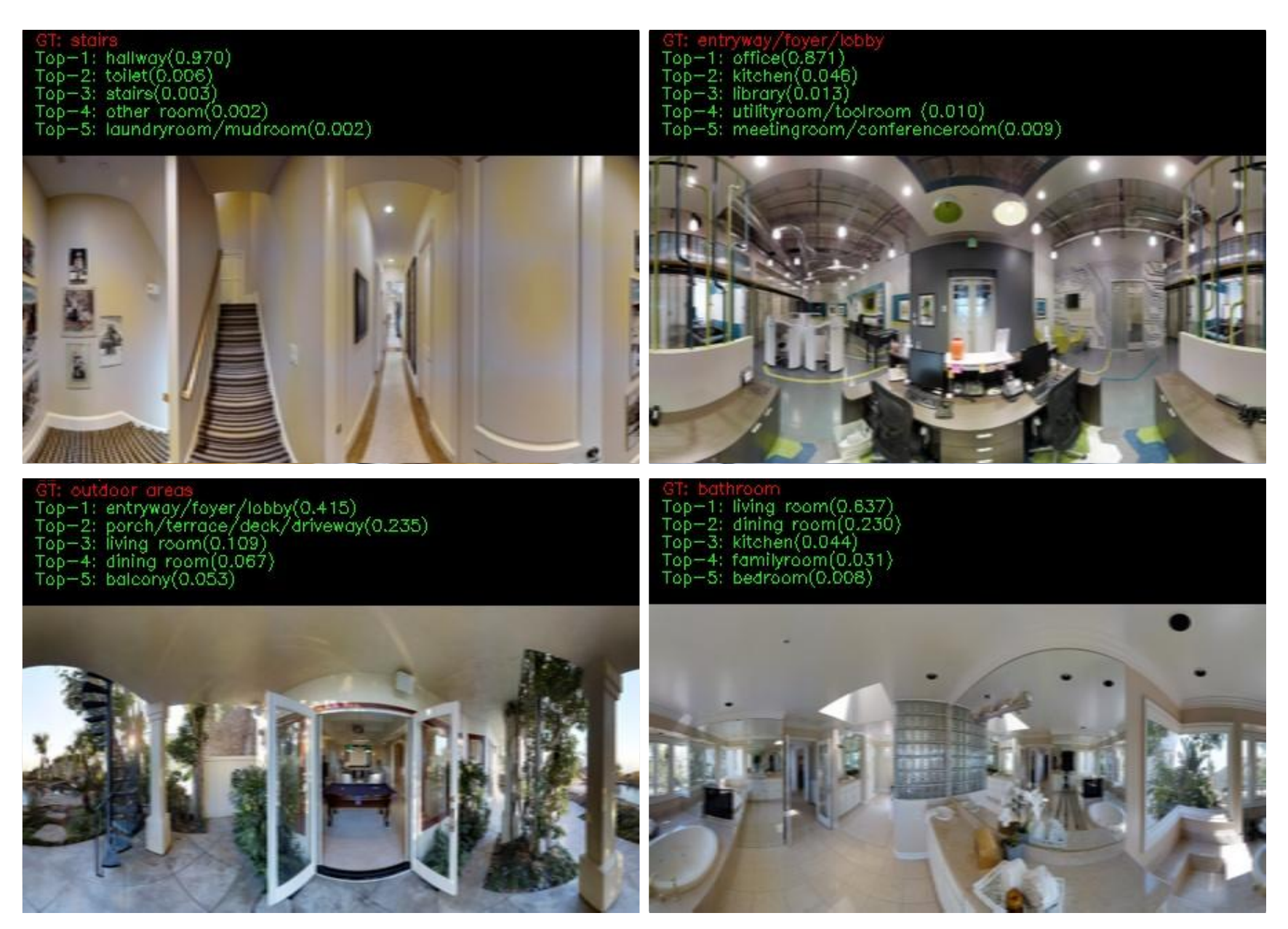}
		\caption{
		    Scene recognition failure examples. The panoramic images have large field-of-view and include several regions in one image. The model thus gets confused.
		}
		\label{supp:fig:sr_failure}
\end{figure}
\begin{figure}[!ht]
        \centering
		\includegraphics[width=0.8\columnwidth]{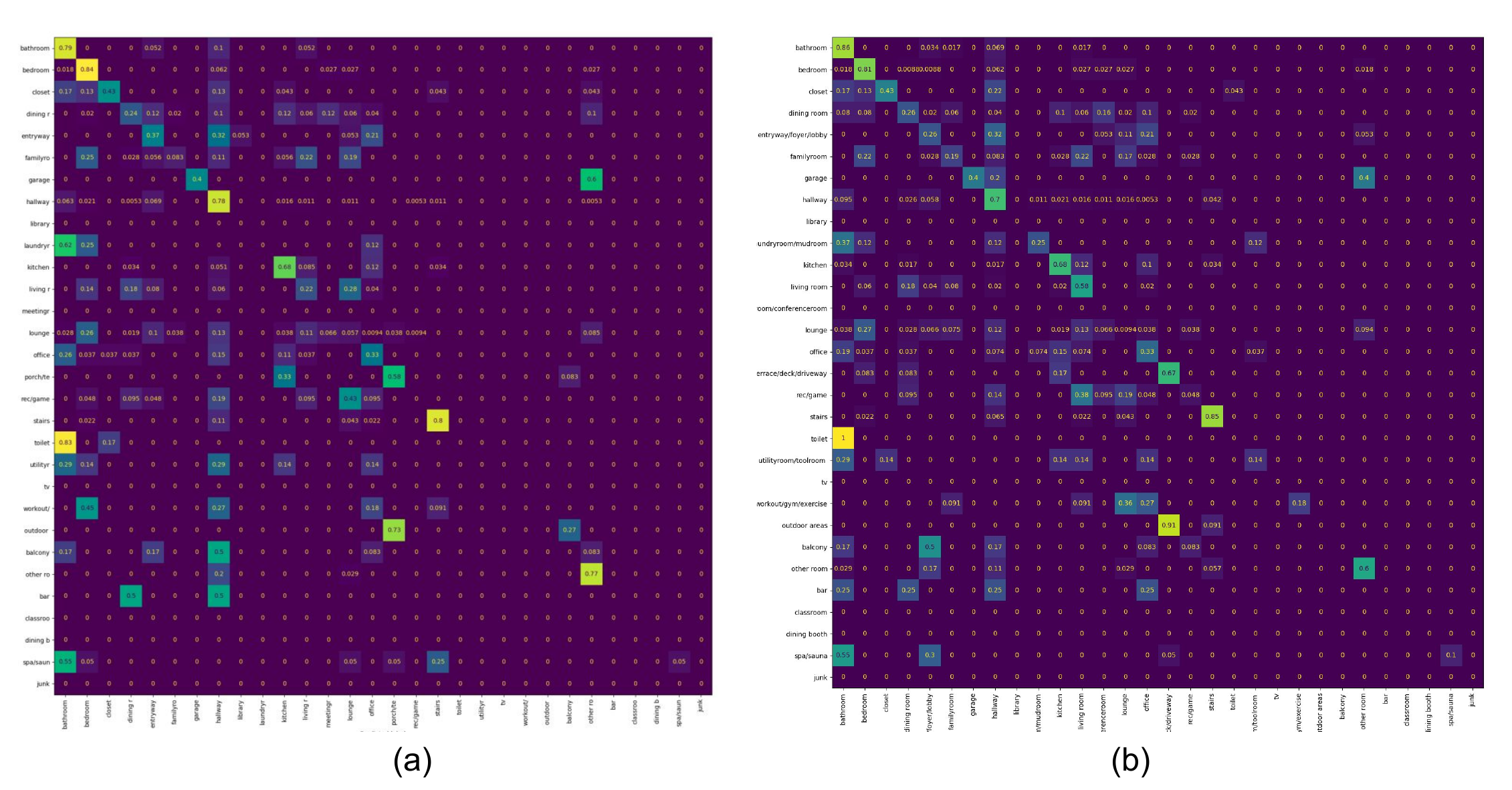}
		\caption{
		Confusion matrix result with the use of different training losses.
		    (a) cross-entropy loss;
		    (b) focal loss.
		The model trained with focal loss shows higher accuracy on infrequent classes.
		}
		\label{supp:fig:perception_conf_mat}
\end{figure}

{\bf Room classification: }
In the main paper, we have presented the average classification accuracy of our trained model.
Here we provide more detailed results and analysis.
\figref{supp:fig:sr_stat} presents bar charts representing various statistics.
The blue bars show the appearance frequency of each class in the test set.
The orange and green bars represent the number of accurately classified samples in each class.
The top-1/5 accuracy is shown above the bars.
This shows that there is a dataset imbalance issue, which we use a focal loss to reduce its influence.

We provide some failure examples of this model in \figref{supp:fig:sr_failure}.
In each example, we provide top-5 predictions and their corresponding probability.
As the input to the model is a panoramic image, which has an extremely large field of view, it can include several regions in one image, which may confuse the model.

We also compare the confusion matrix of the predictions using cross-entropy loss and focal loss. 
The result is shown in \figref{supp:fig:perception_conf_mat}.
The comparison shows that focal loss relieves the imbalance issue existing in the dataset.

\section{More Detail of \textit{Prediction}} \label{supp:sec:prediction}
In this stage, we aim to predict a trajectory based on the observed room-level graph and destination instruction.
A conditional graph generation network is proposed for this purpose. 
We provide more details about the dataset, network, implementation, and results in this section.

\subsection{Dataset}\label{supp:sec:prediction_dataset}

\begin{figure*}[!ht]
        \centering
		\includegraphics[width=1\textwidth]{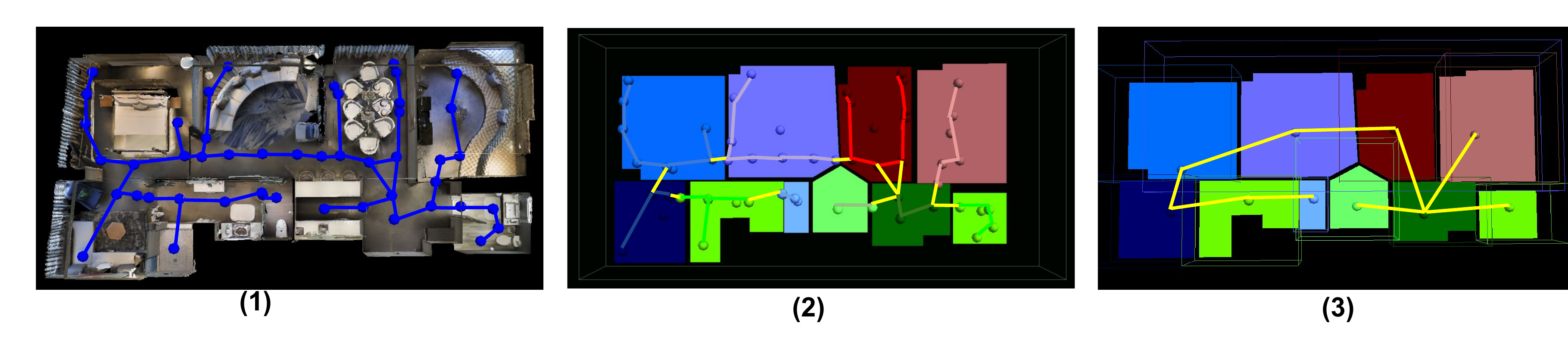}
		\caption{Building region connectivity graph from panorama connectivity graph. (1) Panorama connectivity graph; (2) intra/inter-panorama connectivity graph overlaid on 2D region segmentation map; (3) Region connectivity graph overlaid on 2D region segmentation map
		}
		\label{supp:fig:build_region_graph}
\end{figure*}

\begin{figure*}[!ht]
        \centering
		\includegraphics[width=0.8\textwidth]{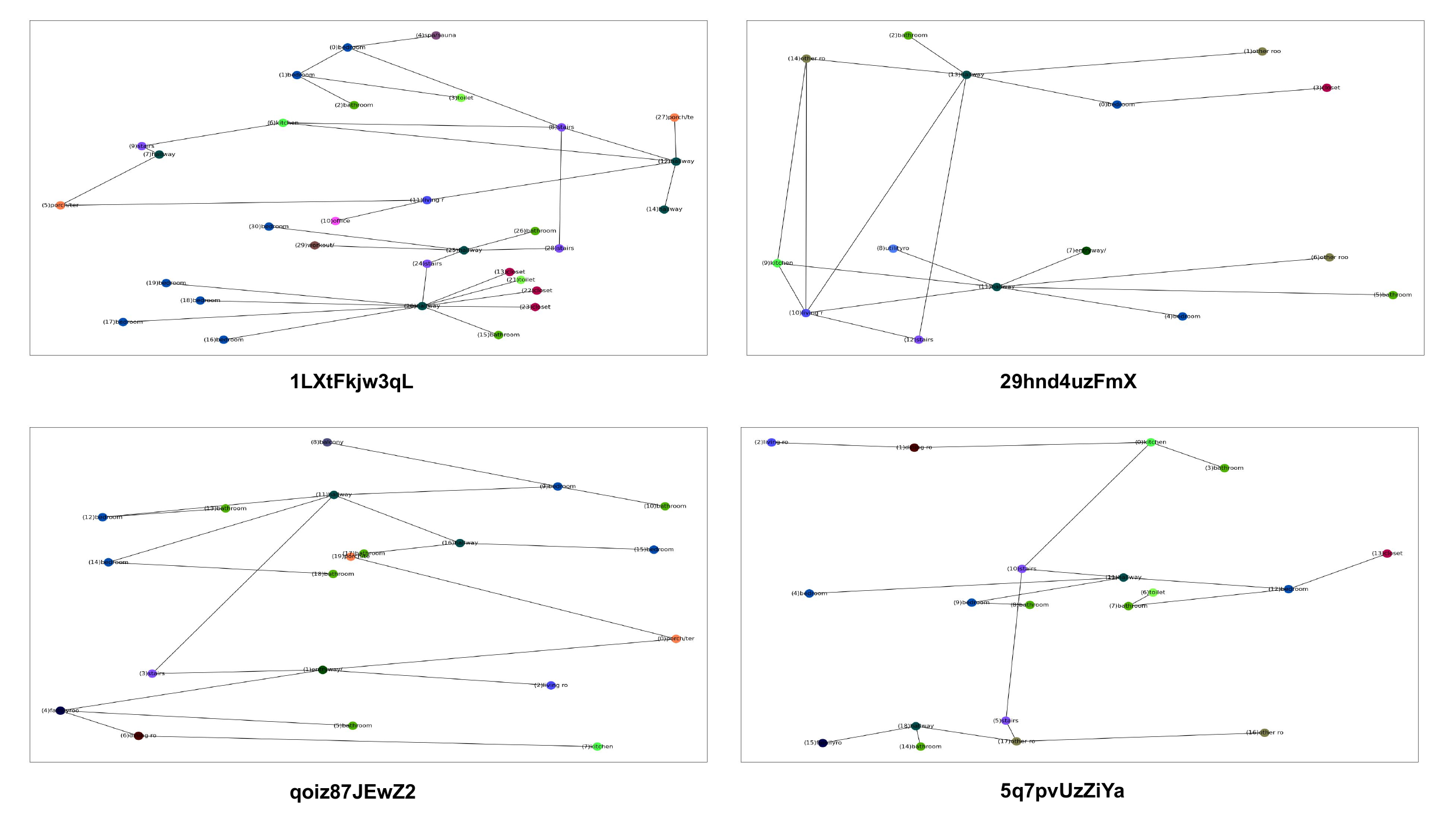}
		\caption{Region connectivity graph examples. Zoom-in for better view.
		}
		\label{supp:fig:region_graph_eg}
\end{figure*}




%
{\bf Room connectivity graphs: }
To create the training data for this task, we first generate room connectivity graphs.
We use \figref{supp:fig:build_region_graph} to illustrate the procedure.
\cite{mattersim} constructs a panorama connectivity graph (\figref{supp:fig:build_region_graph} (1)) by ray-tracing between viewpoints in the Matterport3D scene mesh.
Since Matterport3D provides rich semantic information, including semantic region segmentation (\figref{supp:fig:build_region_graph} (2)), we can group the panoramas according to the rooms they belong to.
The connectivity between rooms can be determined by checking the connectivity between panoramas from different regions, \ie yellow edges in (\figref{supp:fig:build_region_graph} (2)).
Finally, a fused room connectivity graph is formed by fusing yellow edges, \ie removing duplicated edges connecting the same region pairs.
Some graph examples are shown in \figref{supp:fig:region_graph_eg}.

{\bf Dataset construction for training: }
During each training step, we sample a start node and end node from the room connectivity graphs to create a training sample.
Then, all simple trajectories (without loops) that connect the start node and the end node are computed, 
in which we sample one of them as the training sample.
To define the observation and target graph in the trajectory, 
we further sample a connected sub-graph of this trajectory as the observation and keep the remaining as the target graph to be predicted.
We set $B$ as 5 in our experiments, \ie maximum 5 nodes will be generated at each generation step.
For target graphs with less than $B$ nodes, we only compute the loss for the valid nodes.

\subsection{Conditional Graph Generation Network}
\begin{figure}[th!]
        \centering
		\includegraphics[width=0.5\columnwidth]{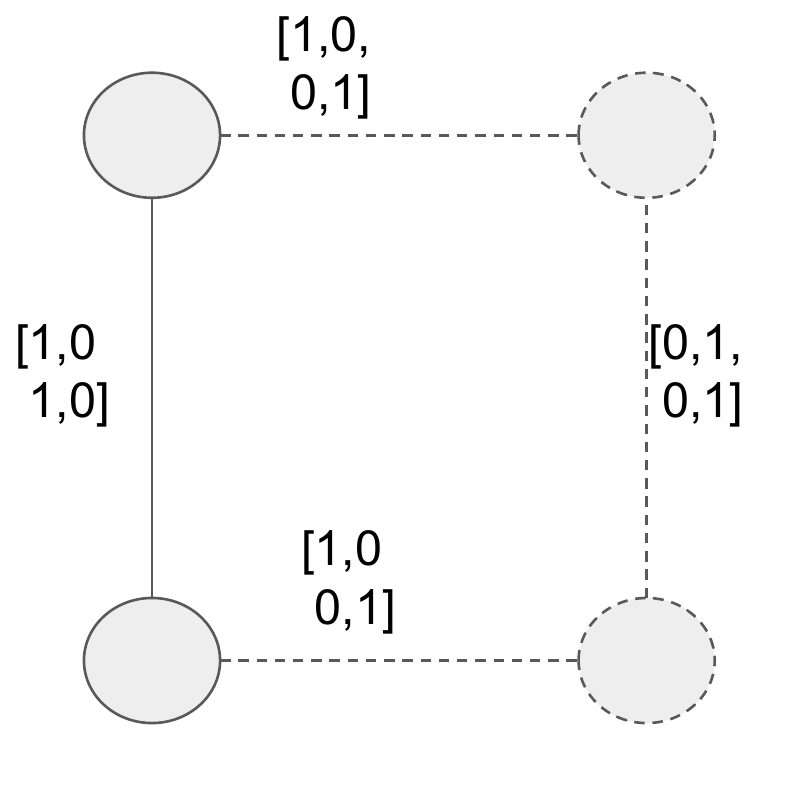}
		\caption{
		    Edge feature initialization by denoting the involvement of existing/new nodes.
		    New nodes are shown as dashed lined circles and edges involving new nodes are shown as dash lines.
		    The first two entries of the edge feature indicate the nature (existing/new) of the first node in the pair while the last two entries indicate that of the second node.
		}
		\label{supp:fig:edge_feat}
\end{figure}

We predict the trajectories using a conditional graph generation network (CGGN).
The inputs to the CGGN are an observation graph and a destination instruction.
We first embed the node features using linear mapping with hidden dimension 256.
The embedding is used as the node initialization.
For the edge feature initialization, we use a vector to denote the involvement of new/existing nodes; see the example in \figref{supp:fig:edge_feat}.
We follow \cite{liao2019gran} to adopt a graph neural network (GNN) for message passing in the graph once node and edges are initialized.
$5$ GNN layers are used in our final model.
The hidden feature size is 256.
In our task, as we need to predict both edges and node feature, \ie region class probability, we create a 3-layer fully-connected network with ReLU activation for node classification. The hidden layer dimension is 128. 

We use Adam as the optimizer, with a batch size of 60 and a learning rate of $10^-3$.
Our best model is trained for 10,000 iterations.

\subsection{Qualitative examples}

\begin{figure}[t]
        \centering
		\includegraphics[width=1\columnwidth]{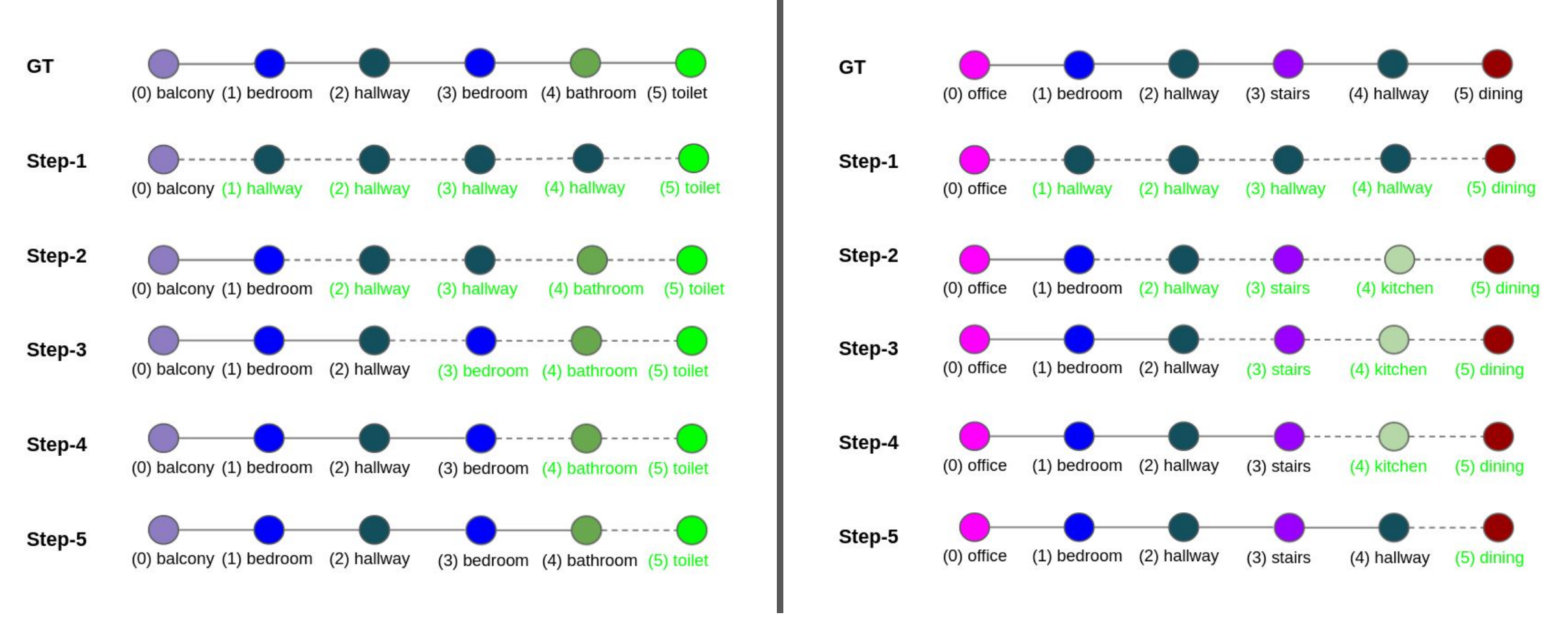}
		\caption{
		    Graph prediction examples. 
		    \textbf{Node: } Description in green means predicted nodes. Only the class with highest probability is shown above for visualization purpose.
		    \textbf{Edge:} Dash lines means predicted edges.
		}
		\label{supp:fig:graph_pred_eg}
\end{figure}

In \figref{supp:fig:graph_pred_eg}, we present two qualitative examples. 
We pre-define some routes and show the graph prediction results at each step.

\section{Navigation: REVERIE} \label{supp:sec:reverie}
\subsection{More experiment and results}
\begin{figure}[t!]
        \centering
		\includegraphics[width=1\columnwidth]{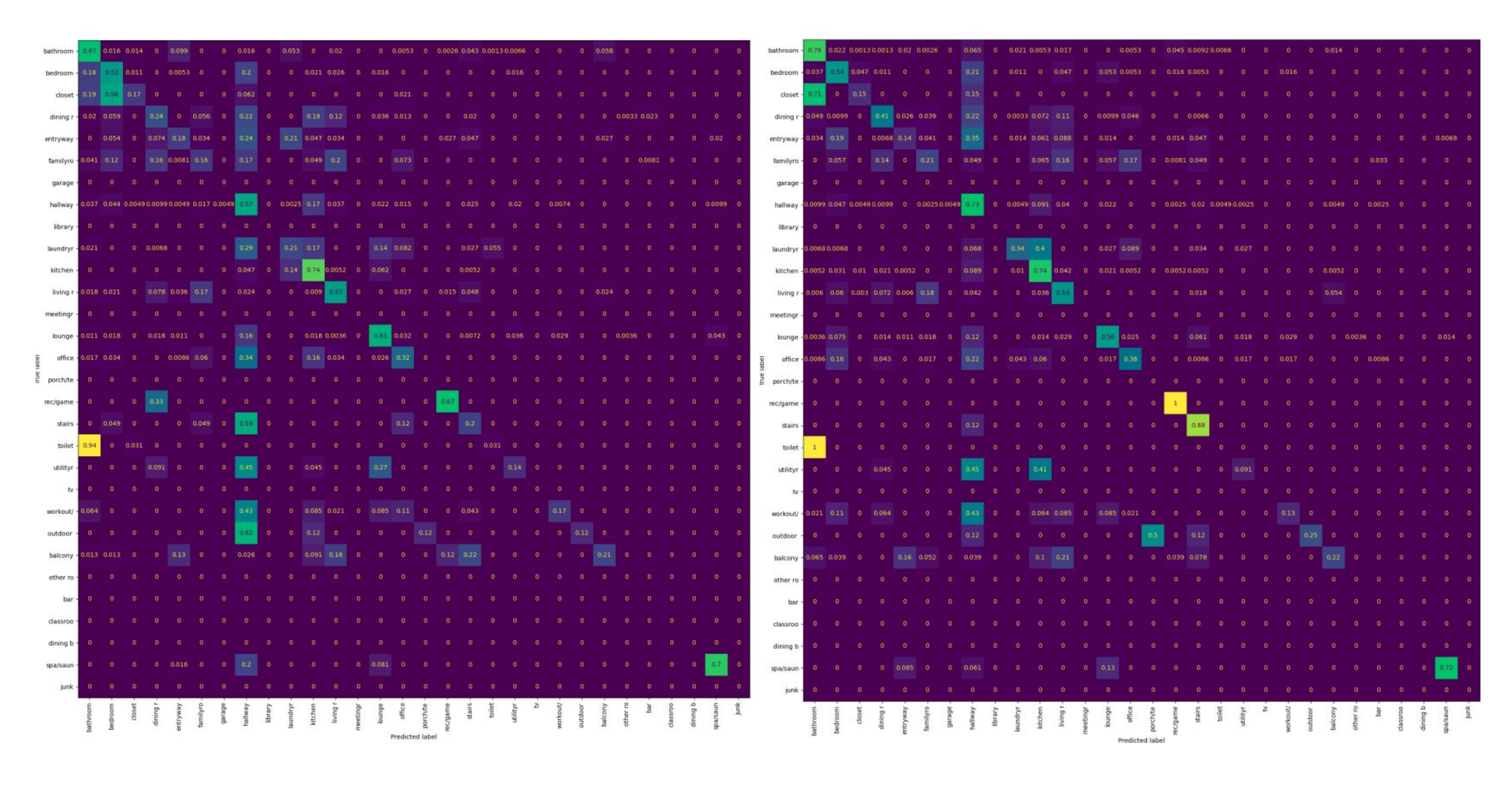}
		\caption{
		Confusion matrix of navigation result in Val-Unseen. (a) Baseline. (b) Ours.
		}
		\label{fig:nav_conf_mat}
		\vspace{-8pt}
\end{figure}

{\bf Stop location confusion matrix: }
We analyse the stop locations of the navigation agents, including \vlnbert Baseline and Ours.
As mentioned in the main paper, we find that 52\% of the failure cases of the Baseline model end up at a location that is not the room type of the goal, namely wrong-room-rate.
We show that the wrong-room-rate is reduced to 42\% when Perception-informed exploration is enabled.
This error rate can be further reduced if a perfect Perception module ($[\textbf{G}, \textbf{P}^*, \textbf{M}]$) is provided.
We illustrate the result using a confusion matrix as shown in \figref{fig:nav_conf_mat}.

\begin{figure*}[t!]
        \subfloat{
            \includegraphics[width=0.48\textwidth]{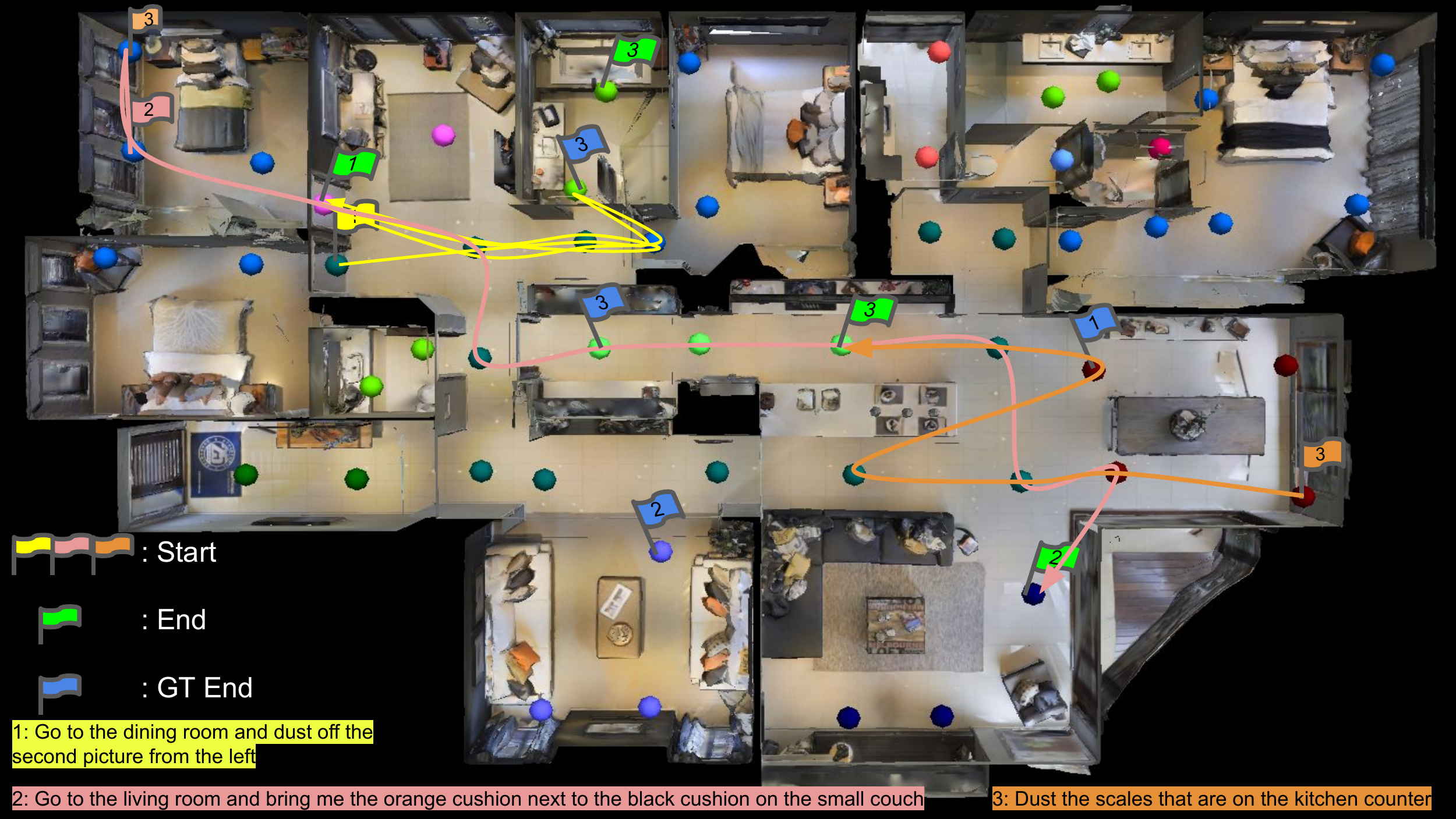}	
            \label{fig:nav_eg_baseline3}}
        \hfill
        \subfloat{
            \includegraphics[width=0.48\textwidth]{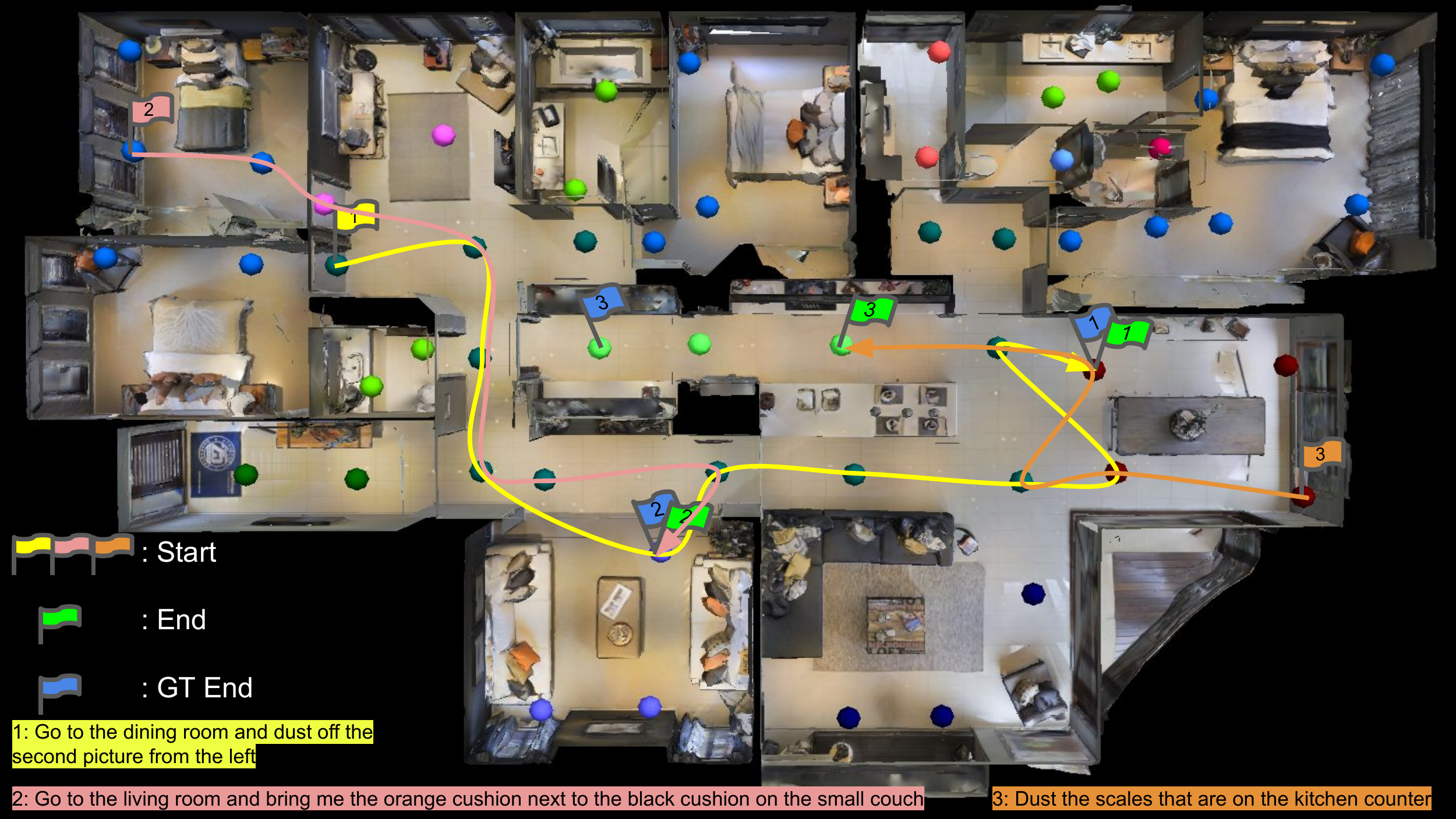}	
            \label{fig:nav_eg_ours3}}

        \subfloat{
            \includegraphics[width=0.48\textwidth]{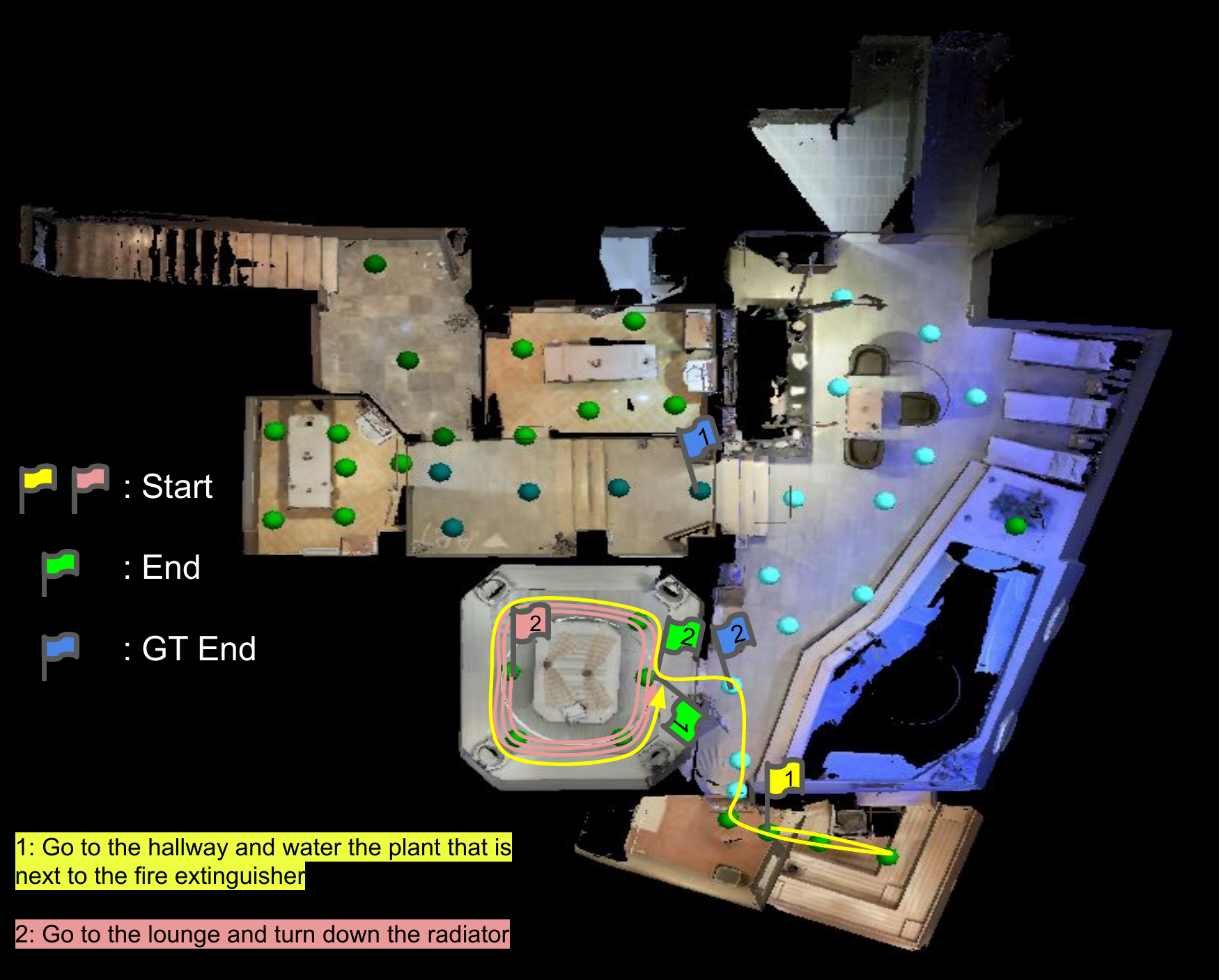}	
            \label{fig:nav_eg_baseline4}}
        \hfill
        \subfloat{
            \includegraphics[width=0.48\textwidth]{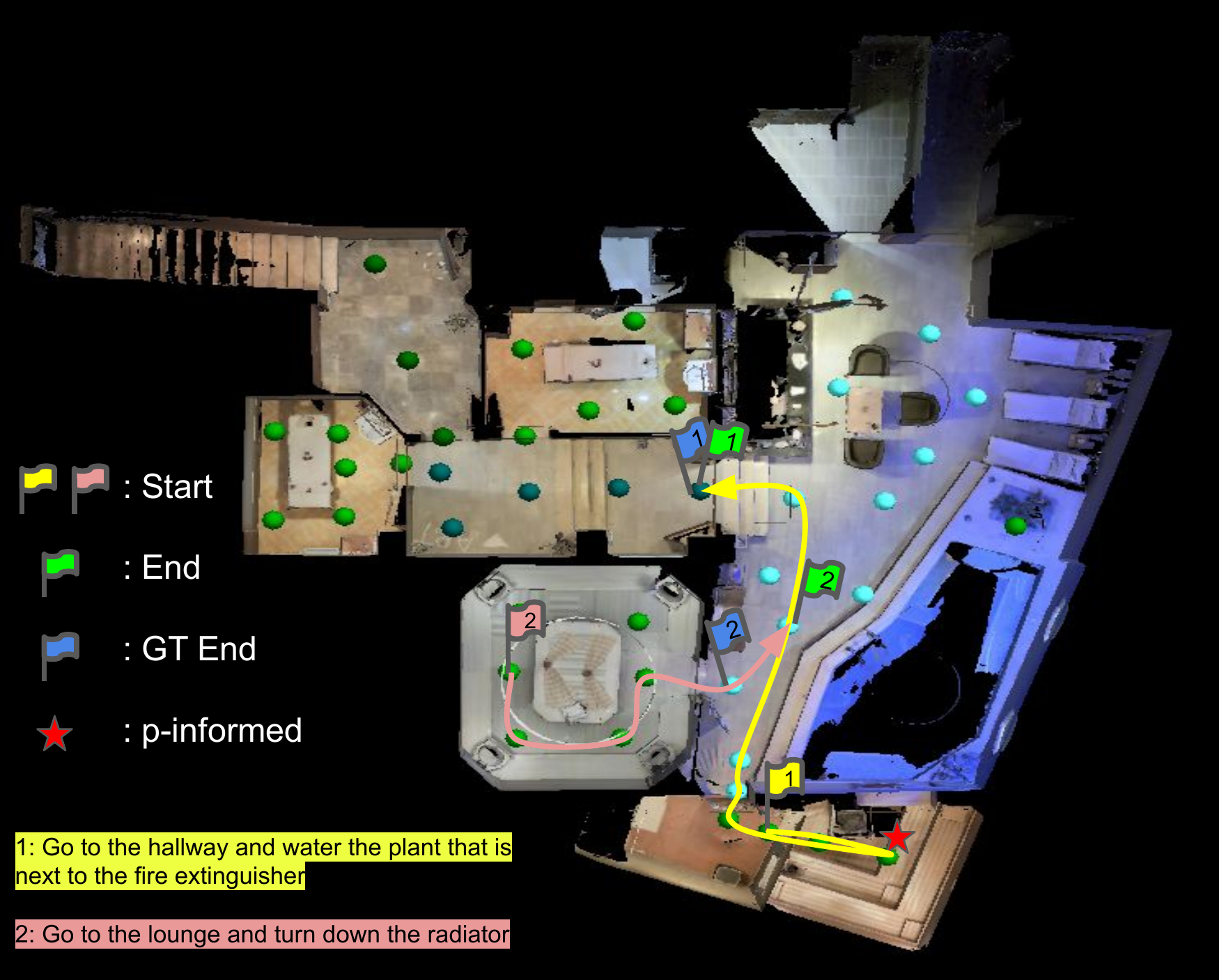}	
            \label{fig:nav_eg_ours4}}

        \subfloat{
            \includegraphics[width=0.48\textwidth]{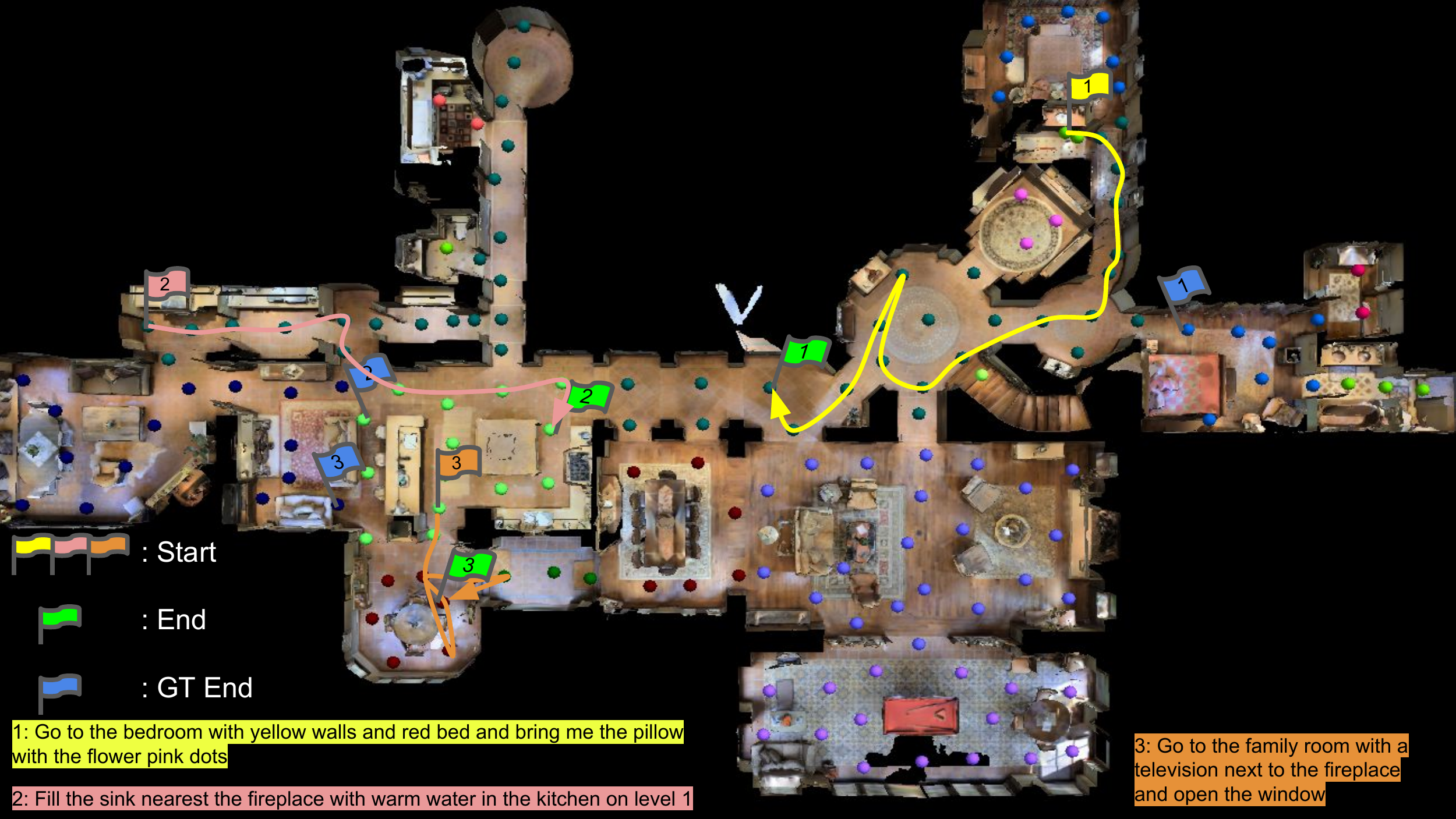}	
            \label{fig:nav_eg_baseline2}}
        \hfill
        \subfloat{
            \includegraphics[width=0.48\textwidth]{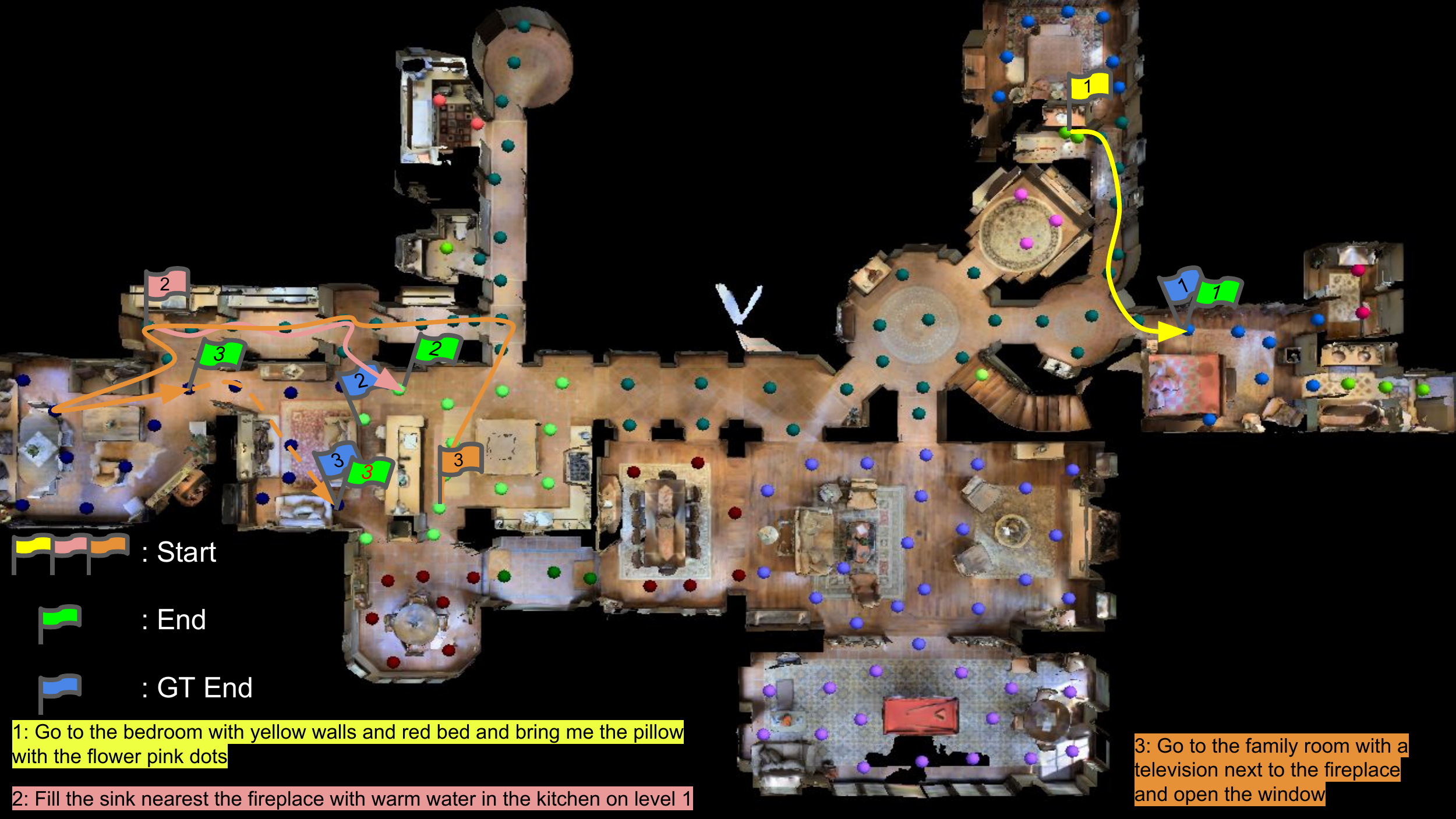}	
            \label{fig:nav_eg_ours2}}
		\caption{
		    More instructed navigation examples are shown above. 
		    The flags are labelled according to the example number.
		    The colored circles represent panoramic viewpoints in different regions.
		    (Left) \vlnbert Baseline
		    (Right) Ours ([$\textbf{G}$, $\textbf{P}$, $\textbf{M}$]-informed).
		    \textbf{Row 1: } we show more navigation examples where our model succeed but Baseline fails.
		    \textbf{Row 2: } 
		    (trajectory-1) We show one example that perception-informed exploration of our model is triggered and prevent early stop, denoted with star sign.
		    (trajectory-2) We show one example that Baseline is trapped in a loop  which never happens to our model due to the map-informed exploration.
		    \textbf{Row 3: } more navigation examples. 
		    In trajectory-3, Ours (15 steps model) stops before reaching the destination. 
		    However, the agent is still able to reach the goal with more steps (maximum 50 steps) (dash line).
		}
		\label{supp:fig:nav_eg}
\end{figure*}

{\bf More qualitative results: }
We provide more navigation examples in \figref{supp:fig:nav_eg} including some of the failure cases of our model.

\section{Navigation: Area-Goal} \label{supp:sec:area_goal}
The key element to our framework is that both the current 3D Scene Graph and predicted graph are fed into the module;
this notion is independent of the exact implementation of the module and can be adapted to a wide variety of navigation tasks.
In the main paper, we have already shown an application example on REVERIE task.
In this section, we will showcase a different navigation task, namely Area-Goal, with the employment of our framework.


  

	

\subsection{Problem Statement and Environment Setting}
Anderson \etal introduce Matterport3D Simulator\cite{mattersim} and the Room-to-Room (R2R) navigation task, which are specifically designed for navigation between panoramic viewpoints in Matterport3D Dataset.
In this Area-Goal task, we focus on high-level topological navigation.
Area-Goal is defined as navigating in a topological map to reach a destination region given abstract instruction about the destination region, which can be specified in a higher level of abstraction, \eg ``go to kitchen" as the instruction.
It does not assume any prior knowledge about the environment structure is known to either the agent or instructor.

As discussed in \secref{supp:sec:prediction_dataset}, we have generated room connectivity graphs for training Prediction module.
The room connectivity graphs can be used for this high-level topological navigation task.
The agent is trained to reach the destination by a series of actions of choosing navigable rooms.

\subsection{Method}
\begin{figure*}[t!]
        \centering
        \includegraphics[width=0.7\textwidth]{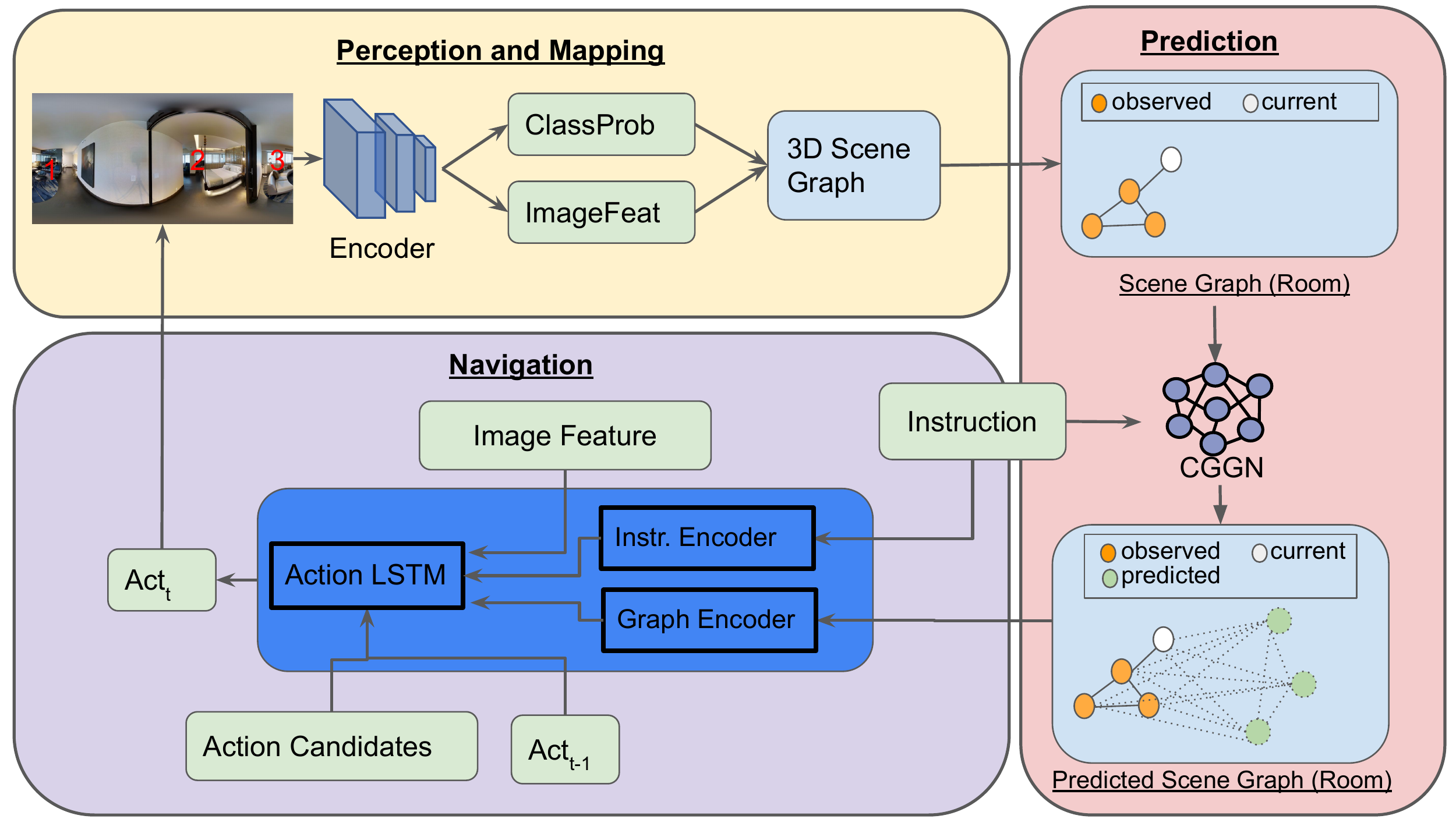}	
		\caption{
		    Schematic of the navigation framework for Area-Goal.
		    A simpler VLN model is implemented and tested in this task.
		}
		\label{fig:area_goal_framework_overview}
		\vspace{-10pt}
\end{figure*}
%
%
We employ the same framework (\figref{fig:area_goal_framework_overview}) in general for this task.
The major difference between Area-Goal model and REVERIE model sits in the implementation of the VLN Model.
In the main paper, we have showcased the application on \vlnbert.
In this Area-Goal task, we aim to show how the Prediction module can help a simple VLN model.
Therefore, in this section, we describe a visual navigation model, which leverages a graph prior predicted by the Prediction module.
Similar to \cite{mattersim}, we model the agent with an LSTM-based recurrent neural network with a graph-attention mechanism.
At each step \textit{t}, 
the agent receives a set of features to predict a distribution over the next action.
The features include
destination instruction,
current panorama feature,
previous action,
the encoded scene graph,
and a set of navigable region candidates' features (region class probability).

\emph{\bf Instruction encoding: }
After receiving a destination instruction, represented by a one-hot vector of the destination class $d$, 
a 3-layer MLP is used to obtain the initial hidden state, 
$h_0 = \text{MLP}_h(d)$.
$h_0$ is used in the attention mechanism.
Another 3-layer MLP is used to initialize the cell state of LSTM, $c_0 = \text{MLP}_c(d)$.
The hidden feature dimension is 256 for the two MLPs.

\emph{\bf Graph encoding: }
Differs from the approach introduced in the main paper, which explicitly extracts subgoals from the predicted graph and provides the VLN model with a subgoal, we directly provide the VLN model with an encoded graph using a graph encoder\cite{Nguyen2019UGT}, which is based on a transformer network with a self-attention mechanism.

\emph{\bf Model action space: }
we define our action space as a set of navigable rooms,
[\textit{Current-Room}, \textit{Room-1}, ..., \textit{Room-N}] \cite{fried2018speaker}.
When \textit{Current-Room} is chosen, we end the navigation episode.

\emph{\bf Action prediction with attention: }
To fully utilize the layout graph for action prediction,
we use a graph-attention mechanism based on \cite{bahdanau2014neural} to identify the most relevant parts of the graph.
Note that all $W_i$ in this section represent a linear mapping layer.
At step-\textit{t}, an attention weight for each node 
is computed as 
$\alpha_{t,i} = (W_1h_{t-1})^T v_{t, i} $,
where $v_{t,i}$ is the node-\textit{i} state at time-\textit{t}.
We normalize the attention weights by 
$\alpha_{t,i} = \alpha_{t,i} / \sum_i \alpha_{t,i}$.

Thus, the attentional graph feature is computed as $\hat{h}_{t-1} = \text{tanh}(W_2[\hat{v}_{t}, h_{t-1}])$,
where $\hat{v}_{t} = \sum_i \alpha_{t,i} v_{t, i}$. 
$\hat{h}_{t-1}$ is concatenated with the current panorama feature and the previous action as the input signal $z_{t}$ to LSTM.
The updated hidden state $h_t$ and cell state $c_t$ are estimated as  $h_t, c_t = \text{LSTM}(z_t, \hat{h}_{t-1}, c_{t-1})$.
The logit of an action (region) candidate-$i$
is computed as
$ W_5 (W_3h_{t} \odot W_4 u_{t, i})$, where $u_{t,i}$ is the candidate's region class probability.
The logits are passed through a SoftMax operator to obtain the probability for each action.

\begin{figure}[t]
        \centering
		\includegraphics[width=0.8\columnwidth]{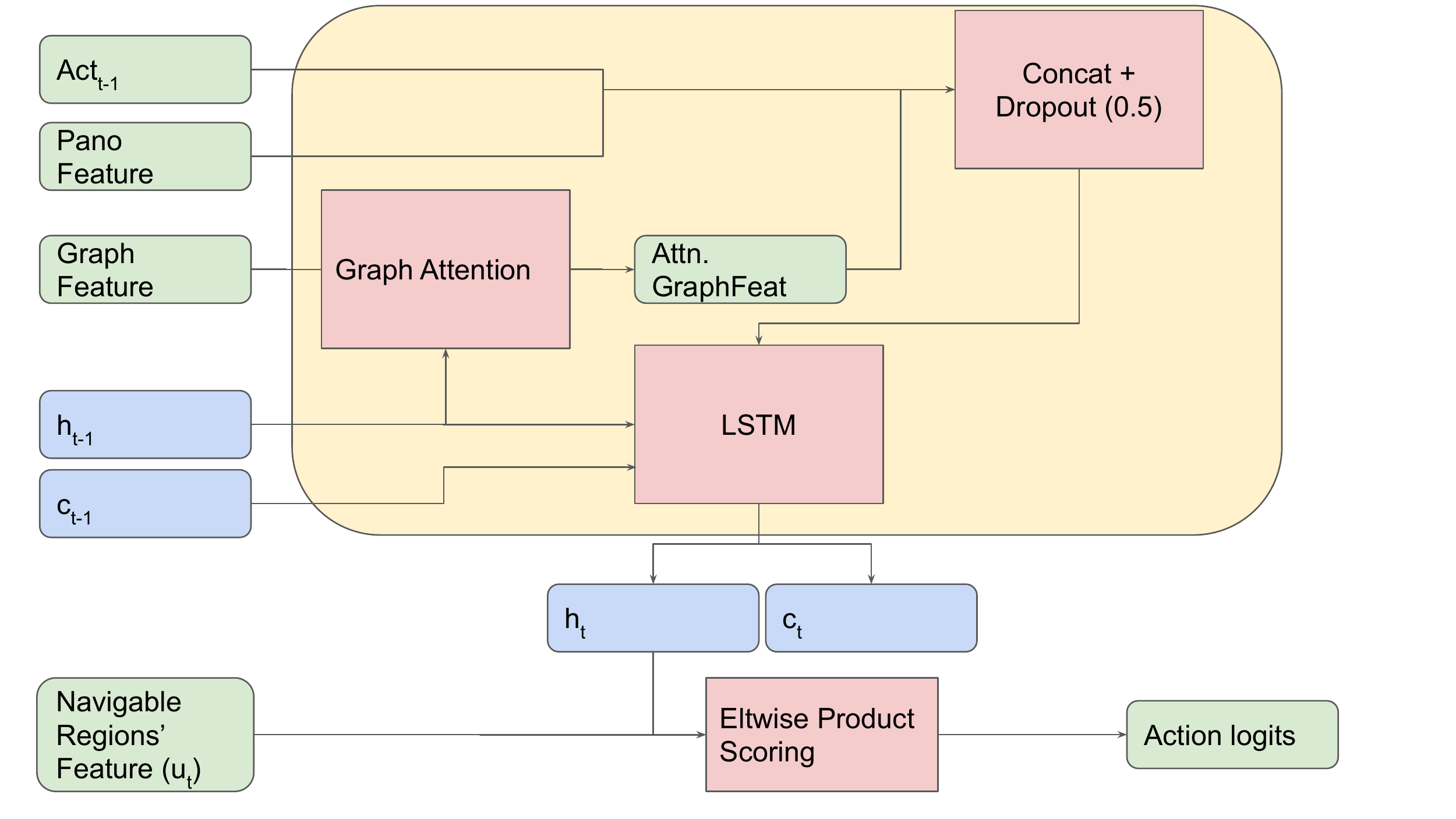}
		\caption{
		    Data flowing in Action LSTM
		}
		\label{supp:fig:action_lstm}
\end{figure}

We provide an illustration of Action LSTM in \figref{supp:fig:action_lstm}. The inputs to the network are detailed below.
\begin{itemize}
    \item $Act_{t-1}$: region class probability of the chosen navigable region in the previous timestep.
    \item $Pano Feature$: panorama image feature extracted from the perception module described above.
    \item $Graph Feature$: graph feature extracted from the graph network described in \cite{Nguyen2019UGT}.
    \item $h_{t-1}, c_{t-1}$: hidden state and cell state of LSTM in the previous timestep.
    \item $Navigable Regions' Feature$: region class probability of the nearby regions
\end{itemize}

%
\begin{table*} [t]  
    \begin{center}
    \resizebox{0.9\textwidth}{!}{%
    \begin{tabular}{ l |  c  c | c c | c c  } 
        & 
        \multicolumn{2}{c|}{\textbf{Val Seen}} & 
        \multicolumn{2}{c|}{\textbf{Val Unseen}} & 
        \multicolumn{2}{c}{\textbf{Test Unseen}} \\
        
        & SPL & Success Rate 
        & SPL & Success Rate 
        & SPL & Success Rate \\
        \hline
        
        \multicolumn{7}{c}{\textbf{Navigation using Perception and/or Prediction}} \\
        \hline
        RANDOM 
        & 0.054 & 17.8 
        & 0.049 & 14.4 
        & 0.039 & 14.0\\
        Baseline (\cite{mattersim}$^\#$)
        & 0.136 & 24.4 
        & 0.086 & 18.3
        & 0.115 & 23.9 \\
        Only Obs. Graph 
        & 0.165 & 29.3
        & 0.099 & 21.4
        & 0.116 & 23.8\\
        Full 
        & \textbf{0.215} & \textbf{35.6 }
        & \textbf{0.102} & \textbf{21.9} 
        & \textbf{0.130} & \textbf{28.8} \\
        \hline
        \hline
        \multicolumn{7}{c}{\textbf{Navigation using Ground Truth Perception}} \\
        \hline
        Baseline$^*$ (\cite{mattersim}$^\#$)
        & 0.156 & 28.8 
        & 0.178 & 38.9
        & 0.179 & 38.3 \\
        \hline
        Full$^*$
        & \textbf{0.269} & \textbf{43.2}
        & \textbf{0.201} & \textbf{40.2}
        & \textbf{0.252} & \textbf{45.9}\\
        \hline
    \end{tabular}
    }
    \end{center}
    \caption{
        Ablation study on navigation result in Area-Goal task.
        ``Baseline ($\#$)": no graph information is used. 
        The agents are similar to the agent proposed in \cite{mattersim} except that we do not provide intermediate instructions to the agent. 
        ``Only Obs. Graph" means that Graph Prediction model is not used. Only observation graph is used as the graph prior for navigator.
        ``$*$": 
        we assume perfect perception for room recognition so the action candidate features are the ground truth room types represented by one-hot vectors.
    }
    \label{table:exp:navigation}
\end{table*}
{\bf Training: }
\cite{mattersim} investigated two training regimes, `teacher-forcing' and `student-forcing'.
For both regimes,
the cross-entropy loss is adopted to maximize the likelihood of the ground-truth target action,
which is defined as 
the next action in the ground-truth shortest-path trajectory from agent's current location to the target location.
`student-forcing' samples the next action from the agent's output probability distribution. 
As this strategy allows the agent to possibly explore the entire environment which could improve the generalization,
we use `student-forcing' regime in our experiments.

\subsection{Experiment and Results}
In the navigation task, a starting position for initializing the simulator and 
a destination region for forming a ground-truth trajectory are required.
For each node in a graph, we pair it with all other nodes to form [start, target] pairs. 
Eventually, we form [41,120, 5,425, 95,115] pairs from training, validation, and test scenes.
To create the train/validation/test splits, we randomly sample 40,000 pairs from the training split to train the navigator, while 1000 from the training split is reserved as `validation\_seen' data.
Another 1000 pairs are sampled from the validation set as `validation\_unseen' data.
Lastly, 5000 pairs from the test set are sampled as `test\_unseen' data.

%
%
{\bf Navigation:}
To evaluate our navigation model, we adopt two commonly used evaluation metrics for navigation \cite{anderson2018evaluation}, Success Rate (SR) and Success weighted by inverse Path Length (SPL).
Success is usually defined as reaching the destination or close enough to the destination in vision-and-language navigation tasks \cite{mattersim, anderson2018evaluation}.
Our task is topological map navigation, so we strictly consider success as reaching the destination.
Since our target is an abstract representation (one-hot vector of a class label), we consider an episode success as long as the agent stops at a place with the destination class label.
For SPL, we dynamically define the ground-truth path as the shortest trajectory connecting the start node and the reached correct destination.
\begin{equation}
\begin{aligned}
    \text{SPL} = \frac{1}{N} \sum_{i=1}^{N} s \frac{l_{s}}{\text{min}(l_{s}, l)}
\end{aligned}
\end{equation}
where $s = 1$ if success else $0$, 
$l_s$ is the shortest ground-truth trajectory length, and $l$ is the length taken by the agent in the episode.
Note that the trajectory length is defined as the graph distance between two nodes instead of the distance between nodes in 3D Cartesian space.

As shown in \tabref{table:exp:navigation}, our RANDOM agent achieves a success rate of about 14.0\% in both validation unseen and test unseen split. 
We implement a baseline model with a sequence-to-sequence agent in which the graph prior is not provided to the navigator.
This simple baseline achieving 23.9\% in the Test set is similar to the approach in \cite{mattersim} except that we do not have intermediate instructions.
To understand the effect of the Perception model, we replace the node class probability with ground-truth, assuming that we have a perfect perception model ``Baseline$^*$". 
With improved perception, the navigator shows a higher success rate and SPL, in all scenes.

To understand how much a graph prior helps navigation, we perform three experiments with various graph priors.
First, we only pass the observation graph prior to the navigation model, and the result shows that an observation graph improves the success rate in most scenes when we compare the result with that of ``Baseline".
Second, we test our full model described above. 
Compared to ``Baseline", we show that learning a graph prediction model increases the navigation success rate and efficiency in both seen environments and unseen environments. 
Third, to understand the limitation/upper bound of our proposed framework, we assume a perfect perception module and graph prediction module in the experiment ``Full$^*$". 
We replace the predictions from the modules with ground-truth class labels and trajectories.
The result suggested that,
(1) an informative graph prior helps navigation if we compare ``Full$^*$" with ``Baseline$^*$".
(2) Although our ``Full" model has shown improvement when compared to the baseline method ``Baseline", there is still a gap between the current model when compared with ``Full$^*$".

In summary, the results suggest that training a navigation agent for topological navigation that can generalize in an unseen environment is challenging, which is consistent with the finding suggested in other visual-language-navigation works.
Nevertheless, we show that learning a probabilistic graph prior is beneficial for navigation tasks.



\newpage


\end{document}